%% file: main.tex
\documentclass[10pt, twocolumn, letterpaper]{article}
\usepackage{kotex}

\usepackage{iccv}
\usepackage[dvipsnames,table]{xcolor}
\definecolor{iccvblue}{rgb}{0.21,0.49,0.74}
\usepackage[pagebackref,breaklinks,colorlinks,allcolors=iccvblue]{hyperref}
\usepackage{graphicx}

\usepackage{amsmath}
\DeclareMathOperator*{\argmin}{arg\,min}
\usepackage{comment}
\usepackage{array}
\newcolumntype{?}[1]{!{\vrule width #1}}
\newcolumntype{I}{!{\vrule}}
\usepackage{multirow}
\usepackage{arydshln}
\title{PoseSyn: Synthesizing Diverse 3D Pose Data from In-the-Wild 2D Data}
\author{
ChangHee Yang${}^{*1}$ \hspace{2em}
Hyeonseop Song${}^{*1}$ \hspace{2em}
Seokhun Choi${}^{*1}$ \\  
Seungwoo Lee${}^{1}$ \hspace{2em}
Jaechul Kim${}^{1}$ \hspace{2em}
Hoseok Do${}^{\dag 1}$ \\
AI Lab, CTO Division, LG Electronics${}^{1}$\\
\tt\small \{changhee.yang, hyeonseop.song, seokhun.choi, seungwoo5.lee, jaechul1220.kim, hoseok.do\} \\\tt\small{@lge.com}}

\begin{document}
\maketitle

\label{abstract}
\begin{abstract}
\input{folder_tex/Abstract_arxiv}
\end{abstract}

\def\thefootnote{*}\footnotetext{These authors contributed equally to this work}%\def\thefootnote{\arabic{footnote}}
\def\thefootnote{\dag}\footnotetext{Corresponding author}%\def\thefootnote{\arabic{footnote}}

\section{Introduction}\label{intro}
\input{folder_tex/Intro_arxiv}

\section{Related Works}\label{relwork}
\input{folder_tex/Related_works_arxiv}

\section{Methods}\label{method}
% \section{Proposed Method}\label{method}
\input{folder_tex/Method_arxiv}

\section{Experiments}\label{Experiments}
\input{folder_tex/Experiments_arxiv}

\section{Conclusion}
\label{Conclusion}
\input{folder_tex/Conclusion_arxiv}

% \clearpage
{
    \small
    \bibliographystyle{ieeenat_fullname}
    \bibliography{main}
}

\clearpage
\appendix
\onecolumn

\begin{center}
\large \textbf{\\PoseSyn: Synthesizing Diverse 3D Pose Data from In-the-Wild 2D Data \\Supplementary Material}
\end{center}

\input{supplementary_main_arixv}

\end{document}

%% file: folder_tex/Abstract_arxiv.tex
Despite considerable efforts to enhance the generalization of 3D pose estimators without costly 3D annotations, existing data augmentation methods struggle in real-world scenarios with diverse human appearances and complex poses.
We propose PoseSyn, a novel data synthesis framework that transforms abundant in-the-wild 2D pose dataset into diverse 3D pose–image pairs.
PoseSyn comprises two key components: Error Extraction Module (EEM), which identifies challenging poses from the 2D pose datasets, and Motion Synthesis Module (MSM), which synthesizes motion sequences around the challenging poses.
Then, by generating realistic 3D training data via a human animation model--aligned with challenging poses and appearances--PoseSyn boosts the accuracy of various 3D pose estimators by up to 14\% across real-world benchmarks including various backgrounds and occlusions, challenging poses, and multi-view scenarios. 
Extensive experiments further confirm that PoseSyn is a scalable and effective approach for improving generalization without relying on expensive 3D annotations, regardless of the pose estimator's model size or design.

%% file: folder_tex/Intro_arxiv.tex
\begin{figure}[t]
\begin{center}
\centerline{\includegraphics[width=0.9\linewidth]{
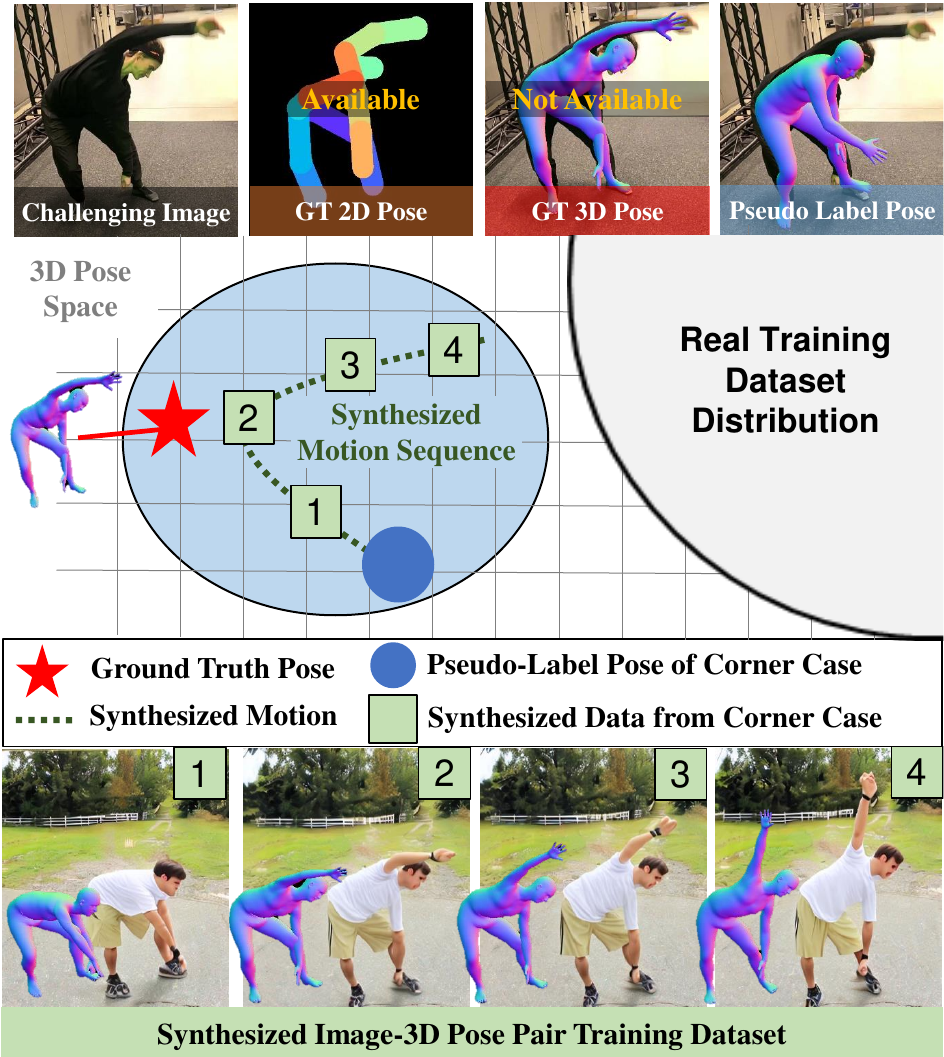}
}
\vskip -0.05in
\caption{
\textbf{PoseSyn's Approach for Corner-Case Pose Generation.}
PoseSyn addresses challenging pose cases without direct costly 3D annotations. Starting from a pseudo-labeled pose (blue circle) that is inaccurate compared to the unknown true 3D pose (red star), PoseSyn generates diverse motion sequences (green dotted lines) around the pseudo-labeled pose to produce new image–pose pairs (numbered 1--4).
PoseSyn effectively bridges the gap between the inaccurate pseudo-label and the desired challenging pose, expanding the training distribution with hard examples.
}
\label{fig:posesyn_concept}
\vskip -0.4in
\end{center}
\end{figure}

Estimating 3D human pose~\cite{shin2023wham,Shetty_2023_CVPR,multi-hmr2024,cha2022multi,yang2023sefd,ge2024humanwild,yang2023sefd,li2022cliff,li2021hybrik, choi2022learning,goel2023humans,kolotouros2019spin,kanazawaHMR18} from RGB images has been extensively researched due to its applications like action recognition, virtual reality, and sports analytics.
Traditional methods rely heavily on large-scale datasets with accurate 3D pose ground truth (GT), but acquiring such 3D annotations often demands multi-camera systems~\cite{cai2022humman, h36m_pami, mehta2018single} or motion capture setups~\cite{vonMarcard2018,kaufmann2023emdb}, which are costly and limited to controlled indoor spaces~\cite{Joo_2017_TPAMI}.
Consequently, existing 3D pose datasets are mainly biased toward laboratory settings, which hinder them from covering in-the-wild challenging poses and reflecting real-world visual elements such as appearance, background, and lighting.
The reliance on indoor data also introduces a domain gap, limiting model robustness in outdoor or uncontrolled environments, thus hindering generalization to real-world scenarios.

To address this gap, prior work has explored data augmentation techniques.
Transformation-based methods~\cite{gong2021poseaug, Gholami_2022_CVPR}, often powered by Generative Adversarial Networks (GANs)~\cite{goodfellow2020generative}, generate new skeletal poses via geometric transformations.
Yet these methods are restricted to modifying keypoints without image-level context.
In contrast, synthetic approaches like PoseGen~\cite{gholami2024posegen} leverage Neural Radiance Fields (NeRF)~\cite{mildenhall2021nerf, su2021nerf} to render novel views, but they often fail to capture the real-world human appearances and backgrounds--limiting their impact on generalization.
Abundant in-the-wild 2D pose datasets offer a cost-effective alternative, but lack corresponding 3D annotations.

We propose PoseSyn, a novel framework that synthesizes diverse 3D pose-image pair data from in-the-wild 2D pose-image pair data to improve the generalization of 3D pose estimators, targeting hard, underrepresented samples.
The key insight is that generating plausible variant data that approximates a challenging pose, then retraining the Target Pose Estimator (TPE) with these variants, improves generalization in real-world scenarios.
Leveraging abundant 2D pose annotations as seeds, PoseSyn generates diverse 3D training samples tailored to each TPE's challenging cases.
Our Error Extraction Module (EEM) identifies challenging images where TPE underperforms, guided by 2D ground truth.
However, GT 3D pose of the challenging image is not available for direct annotation, while the pseudo-labeled 3D pose is often inaccurate.
Naively relying on this inaccurate pose or high-level text description extracted from the challenging image, which are approximations of the hidden GT 3D poses, fails to capture full complexity of pose configurations, making it difficult to generate variant data for such challenging cases.
To mitigate this, our Motion Synthesis Module (MSM) integrates both text descriptions and the inaccurate pose to synthesize motion sequences near the hidden GT pose.
By representing poses as continuous motion rather than isolated frames, MSM reduces ambiguity and produces a range of plausible pose variations that better approximate the desired 3D GT, as shown in Fig.~\ref{fig:posesyn_concept}.
Finally, these generated poses, closely aligned with challenging images, are synthesized into realistic images via a human animation model~\cite{zhu2024champ}, ensuring well-aligned 3D training data.

Moreover, TPEs can vary in parameter size to meet different hardware constraints and accuracy needs in practical applications.
Through experiments across diverse datasets~\cite{vonMarcard2018, kaufmann2023emdb, johnson2011learning, johnson2010clustered, Jhuang:ICCV:2013, Joo_2017_TPAMI, cai2022humman} and multiple TPE architectures~\cite{li2021hybrik,choi2022learning,goel2023humans}, we demonstrate that PoseSyn is a scalable and effective solution for enhancing 3D pose estimation without relying on costly 3D annotations.

In summary, our contributions include: 
\begin{itemize}
\item We propose PoseSyn, a novel framework that generates diverse 3D pose data from in-the-wild 2D pose datasets to address real-world corner cases.
\item We propose Motion Synthesis Module (MSM), which synthesizes motion sequences by leveraging both textual descriptions and initial pose information, extending the coverage of challenging poses.
\item We show consistent improvements in generalization across various datasets and multiple TPE architectures.
\end{itemize}

%% file: folder_tex/Related_works_arxiv.tex
\begin{figure*}[t]
\begin{center}
\centerline{\includegraphics[width=\linewidth]
{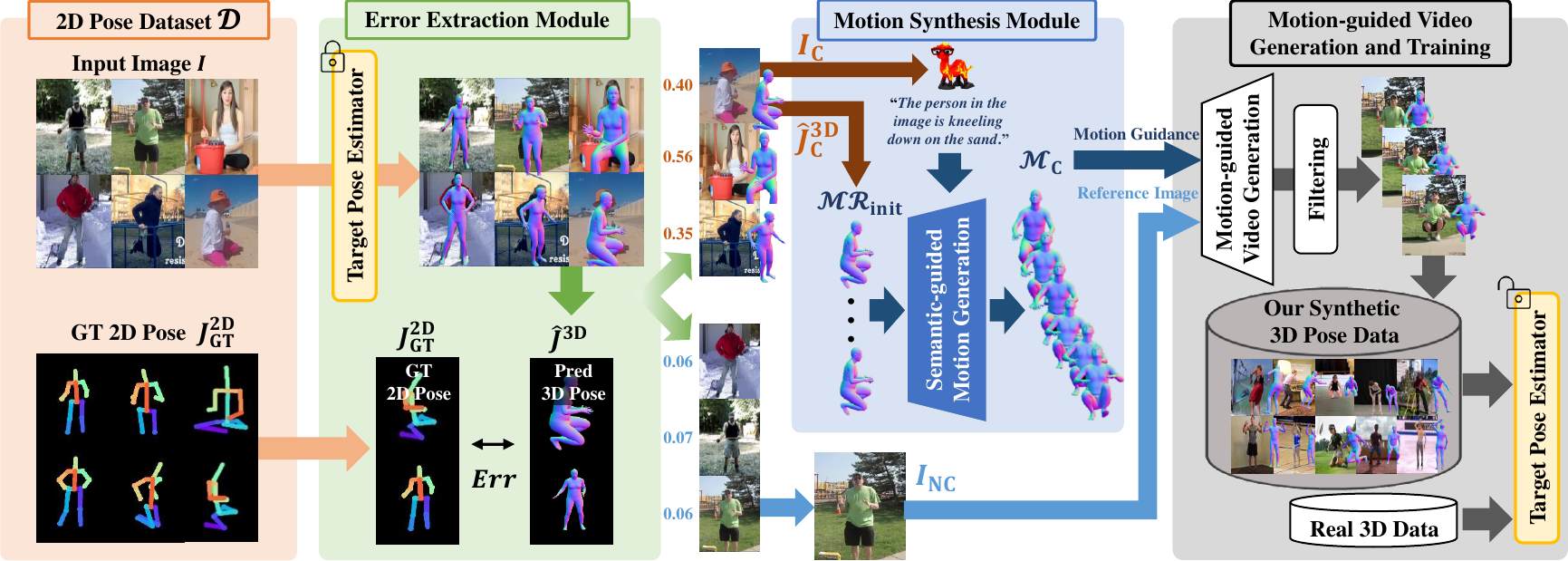}
}
\vskip -0.05in
\caption{
\textbf{Proposed Framework for 3D Pose Data Synthesis.} 
Our framework begins by identifying challenging and non-challenging images in a 2D pose dataset using the Error Extraction Module (EEM). 
EEM isolates poses with high error from the Target Pose Estimator (TPE) as challenging data, which are then processed in the Motion Synthesis Module (MSM) to generate complex motion sequences. 
Subsequently, a motion-guided video generation model creates synthetic training data using non-challenging images as references, which then undergoes a filtering process to finally train the TPE.
}
\label{main_figure}
\end{center}
\vskip -0.35in
\end{figure*}

\subsection{Data Augmentation for Pose Estimation}
Numerous data augmentation techniques have been developed to enhance the generalization performance of 2D and 3D human pose estimation.
Image-level augmentations (\eg, rotation and transformation) diversify training data to some extent~\cite{peng2018jointly,chen2017adversarial,wang2021human,xiao2018simple}, yet they do not drastically change the original image.
Other approaches utilize human parsing, where body parts are segmented and synthesized into real images~\cite{bin2020adversarial,jiang2022posetrans}, but struggle with ambiguities like differentiating front and back views.
Pose-guided or appearance-modifying image generation~\cite{rogez2016mocap,shen2024diversifying,ge20243d,chen2016synthesizing} synthesize images from motion capture or transform human appearance; however, the generated data deviate from real-world distribution as they are not derived from real images.

Other approaches rotate or scale 3D poses and project them into 2D~\cite{li2023cee,gong2021poseaug,Gholami_2022_CVPR}, though mainly limited to 2D-to-3D lifting tasks.
For instance, PoseAug~\cite{gong2021poseaug} and AdaptPose~\cite{Gholami_2022_CVPR} utilize GAN to produce 3D poses via transformations and bone-level augmentations.
More recent work, PoseGen~\cite{gholami2024posegen}, generates diverse poses via GAN and renders images~\cite{su2021nerf}.
However, challenges still remain, as it lacks background and is constrained to predefined human models.
Despite these efforts, the need for augmentation methods that better align with real-world distributions and improve generalization across varied scenarios is evident.

\subsection{Generative Model-based Data Synthesis}
Generative models have recently advanced recognition tasks such as image classification~\cite{hesynthetic, sariyildiz2023fake, qraitem2024fake}, object detection~\cite{wu2023datasetdm, Fang_2024_WACV, kupyn2024dataset}, and segmentation~\cite{nguyen2023dataset, yang2023freemask, fan2024divergen, wu2023datasetdm, jia2024dginstyle, loiseau2024reliability, kupyn2024dataset}.
The core idea is to harness pre-trained generative models~\cite{rombach2022high, zhang2023adding, deepfloyd-if}--which encode rich knowledge from large-scale data~\cite{schuhmann2021laion}--to generate additional training data.
For example, DiverGen~\cite{fan2024divergen} and FreeMask~\cite{yang2023freemask} leverage off-the-shelf text-to-image diffusion models~\cite{rombach2022high,deepfloyd-if,xue2023freestyle} to synthesize diverse training examples with minimal guidance, effectively boosting accuracy on underrepresented classes.
Some approaches use auxiliary modules~\cite{wu2023datasetdm,nguyen2023dataset, Fang_2024_WACV} or light fine-tuning~\cite{rahat2024data, jia2024dginstyle} for task-specific data generation.
A common strategy is to concentrate on corner-cases where standard training data falls short, which is straightforward in object detection and segmentation since these cases can be defined by discrete classes.
For example, if “giraffes” yield low accuracy, one can prompt a generative model for more giraffe images, directly enlarging that corner-case category.

In human pose estimation, however, identifying and synthesizing corner cases is considerably more complex. 
First, the output is a set of continuous keypoint coordinates rather than discrete labels, so corner-case poses cannot be grouped neatly by class. Instead, each problematic pose must be handled at the sample level.
Second, even after identifying the problematic pose, reproducing it via simple text prompts (\eg, “person bending forward with partially occluded right arm”) is nontrivial due to the intricate geometric details in pose.
These challenges highlight a gap in current generative data frameworks, underscoring the need for new strategies to effectively pinpoint and synthesize corner-case poses.

%% file: folder_tex/Method_arxiv.tex
We propose PoseSyn, a novel framework for augmenting in-the-wild 2D pose dataset into 3D human pose data, enabling 3D pose estimators to generalize effectively across in-the-wild scenes.
The framework consists of two main modules: the \textit{Error Extraction Module} (EEM) and the \textit{Motion Synthesis Module} (MSM).
First, EEM identifies challenging poses in the 2D dataset $\mathcal{D}$ by comparing the GT 2D pose $J^{\text{2D}}_{\text{GT}}$ with the 2D projection pose $\hat{J}^{\text{2D}}$ of the pseudo-label 3D pose $\hat{J}^{\text{3D}}$ predicted by a target pose estimator (TPE).
This process produces a set of challenging images and their mis-predicted pose data $\mathcal{D}_{\text{C}} = \{ I_{\text{C}},\hat{J}^{\text{3D}}_{\text{C}} \}$ (Sec.~\ref{sec:EEM}).
Since the actual GT 3D poses for these challenging samples are not available, MSM generates diverse motion sequences to approximate these problematic cases.
To achieve this, MSM first employs a Vision Language Model (VLM)~\cite{wang2024llavallama3} to generate descriptive captions for the challenging images $I_{\text{C}}$.
Then, using a text-to-motion model under the integrated guidance of both caption and the mis-predicted pose, MSM synthesizes challenging motion sequences $\mathcal{M}_{\text{C}}$ (Sec.~\ref{sec:MSM}).
Rather than directly estimating the unknown GT pose as a single pose, these motion sequences serve to extend the pose diversity, covering plausible variations of the challenging scenario.
The synthesized motion sequence $\mathcal{M}_{\text{C}}$ is subsequently utilized in the \textit{Motion-guided Video Generation and Training} stage, where an off-the-shelf human animation model~\cite{zhu2024champ} generates image sequences aligned with $\mathcal{M}_{\text{C}}$. 
Finally, these created data (\ie, images and 3D poses) are filtered, retaining only those that improve generalization, and then used to fine-tune TPE (Sec.~\ref{sec:Training}).
The comprehensive methodology is illustrated in Fig.~\ref{main_figure}.

\subsection{Error Extraction Module}
\label{sec:EEM}
PoseSyn identifies poses within the in-the-wild 2D pose dataset where TPE exhibits high error, termed as challenging poses.
Specifically, let's denote 2D pose dataset as $\mathcal{D} = \{ {I},{J}^{\text{2D}}_{\text{GT}} \} $, where ${I}$ is an image, ${J}^{\text{2D}}_{\text{GT}} \in \mathbb{R}^{N_{\text{2D}} \times 2}$ is the corresponding GT 2D pose, and $N_{\text{2D}}$ is the number of 2D joints.
Through TPE, we predict the pseudo-label 3D pose $\hat{{J}}^{\text{3D}}=\text{TPE }({I})$ and project $\hat{{J}}^{\text{3D}}$ onto 2D image plane $\hat{{J}}^{\text{2D}} = \text{Proj }(\hat{{J}}^{\text{3D}}, f, p)$, where $f$ is the focal length and $p$ is the principal point.
Next, to identify the challenging poses, we calculate the error for each data sample in 2D dataset $\mathcal{D}$:
\begin{equation}
Err = \sum_{n=2}^{N_{\text{2D}}} { \mathbf{w}_n\bigg|\left(\hat{{J}}^{\text{2D},n}-\hat{{J}}^{\text{2D},1}\right) - \left({J}^{\text{2D},n}_{\text{GT}}-{J}^{\text{2D},1}_{\text{GT}}\right)\bigg|},
\end{equation}
where $\hat{{J}}^{\text{2D},n}$ and ${J}^{\text{2D},n}_{\text{GT}}$ are the predicted and GT coordinates of the $n$-th joint, respectively. The pelvis joint is designated as the root joint (\ie, $n=1$) for calculating relative joint positions. The parameter $\mathbf{w}_n$ denotes the weight assigned for the $n$-th joint, which allows important joints to be emphasized when determining challenging poses.
In our method, we assign greater weight to the arms and legs due to their higher variability and complexity in movement, which enhances the accuracy of identifying challenging poses with intricate joint configurations and dynamic orientations.
Based on the metric $Err$, we obtain the top ${K}_{\text{C}}$ poses and their corresponding images as challenging data $\mathcal{D}_{\text{C}} = \{{I}_{\text{C}},\hat{{J}}^{\text{3D}}_{\text{C}} \}$, using the following operation:

\begin{equation}
\mathcal{D}_{\text{C}} = \text{Top}_{{K}_{\text{C}}}\left(Err(d)\;|\; d \in {\mathcal{D}}\right).
\end{equation}

\begin{figure}[t]
\centering
\begin{minipage}[b]{1.0\linewidth}
  \centering
  \centerline{\includegraphics[width=\linewidth]{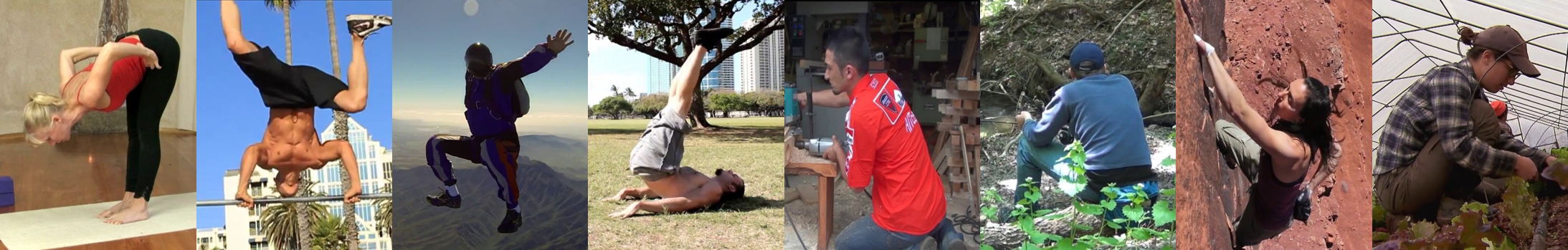}}
  \centerline{\small{(a) Challenging data}}\medskip
\end{minipage}
\vfill
\vskip -0.050in
\begin{minipage}[b]{1.0\linewidth}
  \centering
  \centerline{\includegraphics[width=\linewidth]{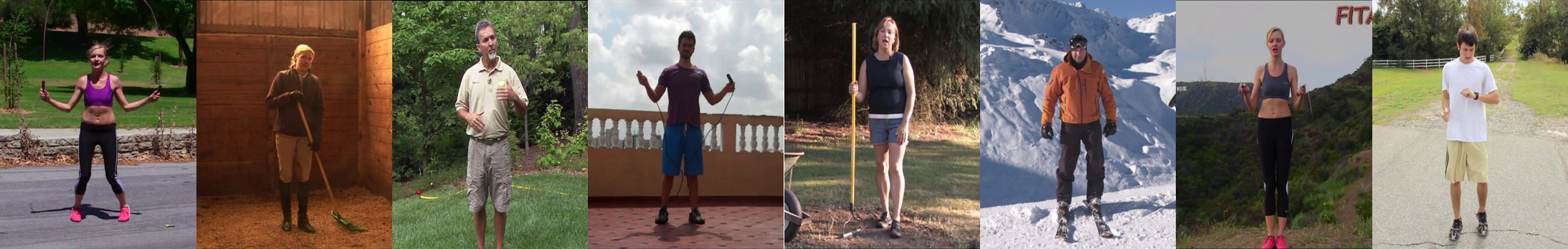}}
  \centerline{\small{(b) Non-challenging data}}\medskip
\end{minipage}
\vskip -0.15in
\caption{
\textbf{EEM Results.} 
The proposed EEM identifies (a) challenging and (b) non-challenging data from in-the-wild 2D pose dataset.
Challenging data includes intricate and dynamic poses, whereas non-challenging data consists of stationary and static poses, demonstrating the effectiveness of the EEM.
}
\label{fig:comp_image}
\vskip -0.1in

\end{figure}

Similarly, we identify non-challenging data $\mathcal{D}_{\text{NC}} = \{{I}_{\text{NC}},\hat{{J}}^{\text{3D}}_{\text{NC}} \}$, which consist of the bottom ${K}_{\text{NC}}$ poses and their corresponding images with respect to the error metric:% $Err$:
\begin{equation}
\mathcal{D}_{\text{NC}} = \text{Top}_{{K}_{\text{NC}}}\left(-Err(d)\;|\;d \in {\mathcal{D}}\right).
\end{equation}
As shown in Fig.~\ref{fig:comp_image}, our EEM isolates the challenging data, typically dynamic and complex poses where TPE struggles. 
The non-challenging data, on the other hand, consists primarily of static poses.
Previous methods~\cite{gholami2024posegen, liu2023poseexaminer} identify challenging poses from unrealistic synthetic or simulated data, thereby lacking the ability to enhance generalization to real-world intricate poses.
However, our EEM identifies the challenging poses in a real-world dataset to create the training data that can improve real-world generalization.

\subsection{Motion Synthesis Module}
\label{sec:MSM}
Given the challenging data $\mathcal{D}_{\text{C}} = \{{I}_{\text{C}},\hat{{J}}^{\text{3D}}_{\text{C}} \}$ identified by EEM, the next necessary step is to generate training data that approximates the actual challenging poses implied by the images, which are unavailable.
One simple approach is to directly utilize the mis-predicted poses $\hat{{J}}^{\text{3D}}_{\text{C}}$.
However, since $\hat{{J}}^{\text{3D}}_{\text{C}}$ deviates from the actual problematic poses, using it alone often results in ineffective training data.
Another straightforward approach is to extract text descriptions from challenging images $I_{\text{C}}$ and generate poses from these descriptions.
Yet, text alone can be ambiguous, making it difficult to precisely capture complex pose configurations.

\begin{figure}[t]
\begin{center}
\centerline{\includegraphics[width=\columnwidth]
{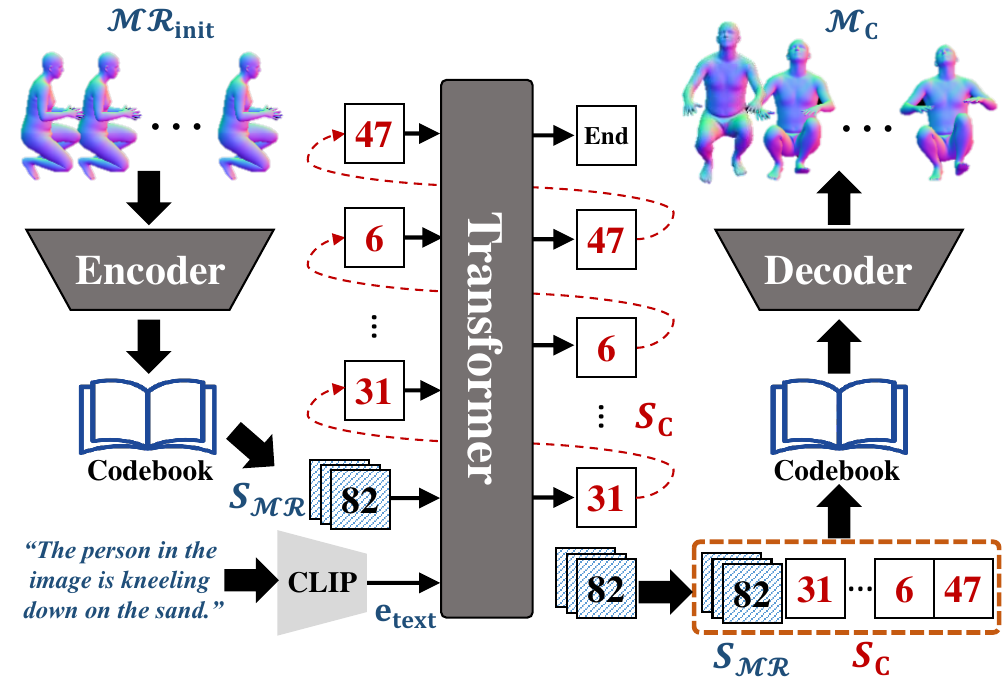}}
\vskip -0.05in
\caption{\textbf{SMG Architecture.}
Proposed Semantic-guided Motion Generation (SMG) augments a mis-predicted pose into motion sequences.
An initial motion representation $\mathcal{MR}_{\text{init}}$ is encoded and mapped into the codebook to produce initial motion indices $\mathcal{S}_{\mathcal{MR}}$.
The transformer takes text embeddings $\mathbf{e}_{\text{text}}$ and $\mathcal{S}_{\mathcal{MR}}$ both as input and generates motion indices for motion sequence $\mathcal{M}_{\text{C}}$.
}
\label{fig:SMG}
\end{center}
\vskip -0.4in
\end{figure}

To overcome these shortcomings, MSM integrates both information in $\mathcal{D}_{\text{C}}$ (\ie, textual cues from the challenging image $I_C$ and the mis-predicted pose $\hat{{J}}^{\text{3D}}_{\text{C}}$) and augments the inaccurate pseudo-label $\hat{{J}}^{\text{3D}}_{\text{C}}$ into motion sequences that encompass plausible pose variations of the challenging scenario.
We instantiate this idea by modifying the existing text-to-motion model, T2M-GPT~\cite{zhang2023generating}, to incorporate both the text description and the mis-predicted pose.
Concretely, a VLM~\cite{wang2024llavallama3} provides text descriptions for each challenging image ${I}_{\text{C}}$ by answering, \textit{``What is the motion of the someone in the image? Please answer similar to \{Text-to-Motion prompt examples\}"} (refer to supplementary material for details).
Next, to address the ambiguity of text-based generation, we introduce an initial motion representation, $\mathcal{MR}_{\text{init}}$, as shown in the SMG architecture in Fig.~\ref{fig:SMG}.
Our empirical findings, presented in Sec.~\ref{sec:effect_pam}, demonstrate that this approach effectively mitigates the ambiguity, surpassing the use of text embedding, $e_{\text{text}}$, alone.

Specifically, we compute $\mathcal{MR}_{\text{init}}=F(\hat{{J}}^{\text{3D}}_{\text{C}} \otimes {T})$ by replicating the pseudo-labeled pose $\hat{{J}}^{\text{3D}}_{\text{C}}$ over ${T}$ time steps and applying a motion representation processing operation $F$~\cite{Guo_2022_CVPR}.
This initial motion representation is then encoded via an encoder $\mathcal{E}$ in Motion VQ-VAE of T2M-GPT: $\mathcal{Z}_{{\mathcal{MR}}}=\mathcal{E}(\mathcal{MR}_{\text{init}})$, where $\mathcal{Z}_{{\mathcal{MR}}}=\{\mathbf{z}_{\mathcal{MR}}^{m}\}^{M}_{m=1}$ are the latent features of initial motion sequences. Here, $M={T}/r$ is a sequence length, and $r$ is a temporal downsampling factor of the encoder.
Each latent feature is mapped into a codebook $\mathcal{C}$ with $Q$ codes $\{\mathbf{c}_q\}^Q_{q=1}$ in the VQ-VAE to obtain initial motion indices $\mathcal{S}_{{\mathcal{MR}}}=\{s_{\mathcal{MR}}^{m}\}^{M}_{m=1}$ as follows:

\begin{equation}
s_{\mathcal{MR}}^m = \argmin_{q}||\mathbf{z}_{\mathcal{MR}}^m-\mathbf{c}_q||_2.
\end{equation}
The initial motion indices $\mathcal{S}_{{\mathcal{MR}}}$ provide additional semantic information for guiding the generation process, helping to complement the insufficient information in the text description alone.
Under the guidance of both textual and initial motion inputs, the motion indices are generated in an autoregressive manner by predicting the distribution of possible next motion indices using a transformer as follows:
\begin{equation}
p(\mathcal{S}_{\text{C}} | \mathbf{e}_{\text{text}},\mathcal{S}_{\mathcal{MR}}) = \prod_{i=0}^{|\mathcal{S}_{\text{C}}|} p(s^i|\mathbf{e}_{\text{text}},\mathcal{S}_{\mathcal{MR}}, s^{<i}).
\end{equation}
Finally, the generated motion indices $\mathcal{S}_{\text{C}}$, along with $\mathcal{S}_{\mathcal{MR}}$, are input into the codebook and decoded through Motion VQ-VAE, producing the augmented challenging motion sequence $\mathcal{M}_{\text{C}}=\{{J}^{\text{3D}}_{\text{C}, l}\}_{l=1}^{L}$ consisting of $L$ 3D joint poses.

\subsection{Motion-guided Video Generation and Training}
\label{sec:Training}

To create images corresponding to the generated motion $\mathcal{M}_{\text{C}}$, we utilize Champ~\cite{zhu2024champ}, an off-the-shelf human animation model that applies motion guidance to reference RGB images.
Using RGB images ${I}_{\text{NC}}$ from non-challenging data $\mathcal{D}_{\text{NC}}$, we generate a human animated video $\mathcal{V}_{\text{C}}=\{I_{\text{C}, l}\}_{l=1}^{L}$ based on the generated motion $\mathcal{M}_{\text{C}}$.
To align the global orientation of $\mathcal{M}_{\text{C}}$ with the human orientation in ${I}_{\text{NC}}$, we pre-process the parameters to ensure natural human animation.
Specifically, among the camera parameters, SMPL pose, and SMPL shape parameters from ${I}_{\text{NC}}$, we replace and adjust the pose parameters with those from generated motion $\mathcal{M}_{\text{C}}$ (refer to supplementary materials for details).
Unlike previous method~\cite{gholami2024posegen}, which utilizes NeRF~\cite{su2021nerf} to render novel-view human images from generated 3D poses--often resulting in artifacts like background absence and limited human appearance diversity--our approach leverages diverse reference images ${I}_{\text{NC}}$.
This approach produces images of varied backgrounds and human appearances, making them more suitable as training data for image-to-3D pose estimators.

While these animated videos exhibit visually appealing qualities that align with the generated motions, artifacts such as the blending between human figures and their backgrounds can still occur.
To filter out noisy samples, each generated image $I_{\text{C}, l} \in \mathcal{V}_{\text{C}}$ is processed by TPE to predict the 3D joint pose
$\hat{{J}}^{\text{3D}}_{l} = \text{TPE}(I_{\text{C}, l})$.
We then compute the error $\text{Err}_{\text{3D},l}$ for each image $I_{\text{C}, l}$ between the predicted 3D joint pose $\hat{{J}}^{\text{3D}, n}_{l}$ and the corresponding generated 3D joint pose ${J}^{\text{3D}}_{\text{C}, l}$ (as derived in Sec.~\ref{sec:MSM}), as follows:

\vskip -0.2in
\begin{equation}
\label{eq:filtering_error}
\text{Err}_{\text{3D},l} = \sum_{n=2}^{N_{\text{3D}}} \left| \left( \hat{{J}}^{\text{3D}, n}_{l} - \hat{{J}}^{\text{3D}, 1}_{l} \right) - \left( {J}^{\text{3D}, n}_{\text{C}, l} - {J}^{\text{3D}, 1}_{\text{C}, l} \right) \right|,
\end{equation}
where ${J}^{\text{3D}, n}_{\text{C}, l}$ and ${J}^{\text{3D}, 1}_{\text{C}, l}$ denote the generated 3D joint pose for the $n$-th joint and root joint, respectively, and $N_{\text{3D}}$ denotes the number of 3D joints.
Any images with $\text{Err}_{3D,l}$ above the threshold $\tau$ are discarded. 
Finally, the remaining high-quality 3D pose data are combined with the original pre-training real data, to re-train TPE, enhancing its in-the-wild generalization capability.

%% file: folder_tex/Experiments_arxiv.tex
\begin{figure*}[ht]
\begin{center}
\centerline{\includegraphics[width=1.0\textwidth]{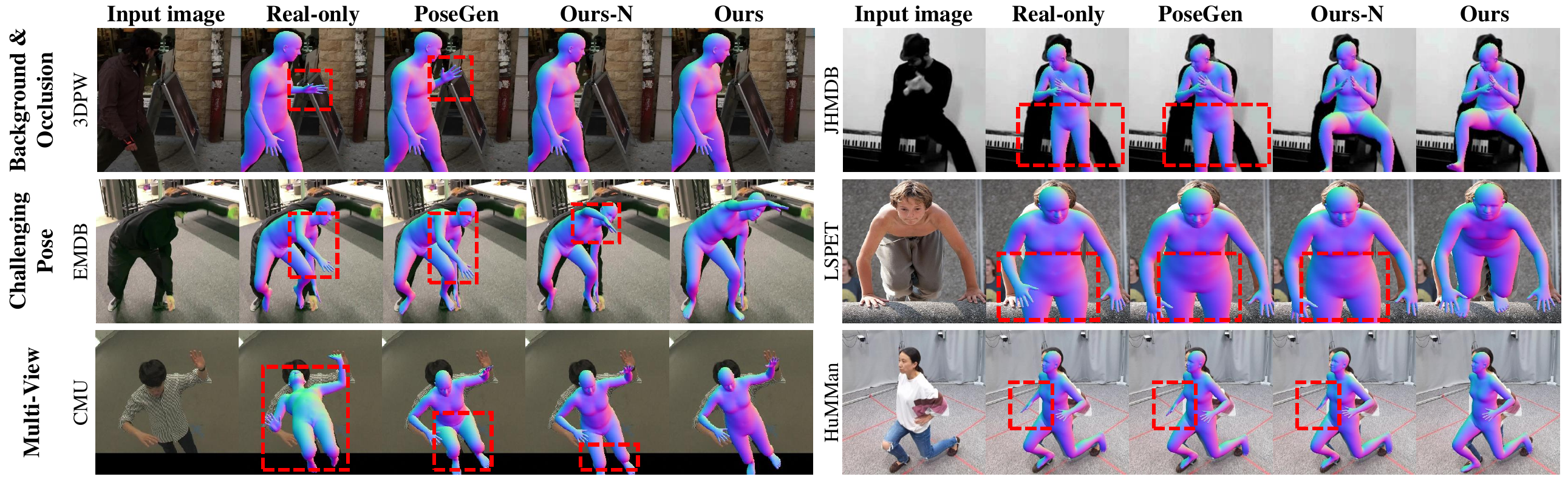}}
\vskip -0.10in
\caption{
\textbf{Qualitative Comparison with Baselines.} 
3DCrowdNet trained with only real data exhibits inaccurate 3D pose predictions with limited generalization capability.
Baselines (\ie, PoseGen and Ours-N) show limited performance gains, but our approach achieves more accurate pose predictions across diverse real-world datasets with various backgrounds and occlusions, challenging poses, and multi-views.
Red boxes highlight areas of incorrect predictions in models trained with real-only data and baseline methods.
}
\label{fig:main_qual_image}
\end{center}
\vskip -0.3in
\end{figure*}

\begin{table*}[]
\centering
\resizebox{\linewidth}{!}{
\begin{tabular}{c|c|cccccccccc|cc||cc|c} 
\hline
\multirow{2}{*}{\rotatebox[origin=c]{90}{TPE}} & \multirow{2}{*}{Method} & \multicolumn{2}{c|}{$\textbf{3DPW}$} & \multicolumn{2}{c|}{$\textbf{EMDB}$} & \multicolumn{2}{c|}{$\textbf{CMU\_171204}$} & \multicolumn{2}{c|}{$\textbf{CMU\_171026}$} & \multicolumn{2}{c|}{$\textbf{HuMMan}$} & \multicolumn{2}{c||}{$\textbf{Mean}$} & \multicolumn{1}{c|}{$\textbf{LSPET}$} & \multicolumn{1}{c|}{$\textbf{JHMDB}$} & \multicolumn{1}{c}{$\textbf{Mean}$} \\ 
\cline{3-17}
 & & \multicolumn{1}{c|}{\scriptsize{$\textbf{MPJPE}\downarrow$}} & \multicolumn{1}{c|}{\scriptsize{$\textbf{PA-MPJPE}\downarrow$}} & \multicolumn{1}{c|}{\scriptsize{$\textbf{MPJPE}\downarrow$}} & \multicolumn{1}{c|}{\scriptsize{$\textbf{PA-MPJPE}\downarrow$}} & \multicolumn{1}{c|}{\scriptsize{$\textbf{MPJPE}\downarrow$}} & \multicolumn{1}{c|}{\scriptsize{$\textbf{PA-MPJPE}\downarrow$}} & \multicolumn{1}{c|}{\scriptsize{$\textbf{MPJPE}\downarrow$}} & \multicolumn{1}{c|}{\scriptsize{$\textbf{PA-MPJPE}\downarrow$}} & \multicolumn{1}{c|}{\scriptsize{$\textbf{MPJPE}\downarrow$}} & \multicolumn{1}{c|}{\scriptsize{$\textbf{PA-MPJPE}\downarrow$}} & \multicolumn{1}{c|}{\scriptsize{$\textbf{MPJPE}\downarrow$}} & \multicolumn{1}{c||}{\scriptsize{$\textbf{PA-MPJPE}\downarrow$}} & \multicolumn{1}{c|}{\scriptsize{\textbf{PCK$^{0.6}_{h}$$\uparrow$}}} & \multicolumn{1}{c|}{\scriptsize{\textbf{PCK$^{0.6}_{h}$$\uparrow$}}} & \multicolumn{1}{c}{\scriptsize{\textbf{PCK$^{0.6}_{h}$$\uparrow$}}} \\ 
\hline \hline
\rowcolor[HTML]{EFEFEF}  \cellcolor[HTML]{FFFFFF} \multirow{4}{*}{\rotatebox[origin=c]{90}{\footnotesize{3DCrowdNet}}} & 
Real-only & 81.7 & 51.1 & 115.8 & 71.2 & 108.8 & 72.5 & 110.7 & 70.4 & 98.9 & 65.8 & 103.2 & 66.2 & 69.5 & 80.5 & 75.0 \\ 
& PoseGen & 80.0 & 50.0 & 113.1 & 70.5 & 104.0 & 68.6 & 106.8 & 68.3 & 94.5 & 64.0 & 99.7 & 64.3 & 70.0 & 79.6 & 74.8 \\ 
& Ours-N & 78.6 & 49.7 & 111.8 & 69.7 & 103.1 & 68.8 & 105.8 & 68.2 & $\textbf{93.1}$ & 63.9 & 98.5 & 64.1 & 70.5 & 81.0 & 75.8 \\ 
& $\textbf{Ours}$ & $\textbf{77.4}$ & $\textbf{48.9}$ & $\textbf{111.0}$ & $\textbf{68.3}$ & $\textbf{101.0}$ & $\textbf{67.3}$ & $\textbf{105.0}$ & $\textbf{67.9}$ & $\textbf{93.1}$ & $\textbf{62.3}$ & $\textbf{97.5}$ & $\textbf{62.9}$ & $\textbf{70.7}$ & $\textbf{81.2}$ & $\textbf{76.0}$ \\ 
\hline
\rowcolor[HTML]{EFEFEF}  \cellcolor[HTML]{FFFFFF} \multirow{4}{*}{\rotatebox[origin=c]{90}{Hybrik}} & 
Real-only & 88.0 & 48.6 & 155.4 & 104.1 & 117.5 & 79.8 & 125.6 & 82.2 & 119.7 & 75.1 & 121.2 & 78.0 & 70.7 & 78.7 & 74.7 \\ 
& PoseGen & 84.8 & 47.6 & 146.9 & 101.0 & 120.8 & 83.2 & 124.8 & 90.9 & 111.2 & 73.3 & 117.9 & 79.6 & 71.1 & 78.7 & 74.9 \\ 
& Ours-N & 81.1 & 46.5 & 141.1 & 99.2 & 115.7 & 80.1 & 127.0 & 86.4 & 100.4 & $\textbf{70.4}$ & 113.1 & 76.5 & 71.3 & 78.7 & 75.0 \\ 
& $\textbf{Ours}$ & $\textbf{78.4}$ & $\textbf{46.2}$ & $\textbf{129.9}$ & $\textbf{90.2}$ & $\textbf{100.3}$ & $\textbf{68.9}$ & $\textbf{119.1}$ & $\textbf{77.1}$ & $\textbf{95.3}$ & 71.1 & $\textbf{104.6}$ & $\textbf{70.7}$ & $\textbf{71.7}$ & $\textbf{78.8}$ & $\textbf{75.3}$ \\ 
\hline
\rowcolor[HTML]{EFEFEF}  \cellcolor[HTML]{FFFFFF} \multirow{4}{*}{\rotatebox[origin=c]{90}{\footnotesize{4DHumans}}} &  
Real-only & 81.3 & 54.3 & 116.3 & 79.1 & 115.1 & 82.0 & 115.3 & 74.2 & 106.1 & 73.8 & 106.8 & 72.7 & 86.3 & 88.3 & 87.3 \\ 
& PoseGen & 81.1 & 54.1 & 114.5 & 78.3 & 111.3 & 79.2 & 114.5 & 71.3 & 105.3 & 73.8 & 105.3 & 71.3 & 86.1 & 88.4 & 87.3 \\ 
& Ours-N & 80.6 & 53.0 & 112.6 & 77.1 & 106.6 & 74.9 & 110.1 & 68.2 & 101.0 & 70.4 & 102.1 & 68.7 & 86.4 & 88.5 & 87.5 \\ 
& $\textbf{Ours}$ & $\textbf{77.0}$ & $\textbf{52.1}$ & $\textbf{108.6}$ & $\textbf{74.8}$ & $\textbf{104.1}$ & $\textbf{72.9}$ & $\textbf{107.8}$ & $\textbf{65.9}$ & $\textbf{98.0}$ & $\textbf{68.5}$ & $\textbf{99.1}$ & $\textbf{66.8}$ & $\textbf{86.6}$ & $\textbf{88.8}$ & $\textbf{87.7}$ \\ 
\hline
\end{tabular}
}
\vskip -0.1in
\caption{\textbf{Quantitative comparison with baselines.} 
Pose estimation performance is evaluated with MPJPE and PA-MPJPE metrics for 3D pose datasets (\ie, 3DPW, EMDB, CMU, and HuMMan), and with PCKh metric for 2D pose datasets (\ie, LSPET and JHMDB). 
For three TPEs (\ie, 3DCrowdNet, Hybrik, and 4DHumans), our approach outperforms other baselines (\ie, PoseGen and Ours-N) across each dataset, demonstrating the highest generalization improvement in real-world scenarios.}
\vskip -0.25in
\label{table1}
\end{table*}

\subsection{3D Pose Estimators and Baselines}
We evaluated three image-to-3D pose estimation models as our target pose estimators (TPEs): Hybrik~\cite{li2021hybrik}, 3DCrowdNet~\cite{choi2022learning}, and 4DHumans~\cite{goel2023humans}.
Hybrik, which bridges the gap between 3D keypoint estimation and body mesh estimation via a hybrid analytical-neural inverse kinematics, is widely used for its strong performance in 3D human pose estimation.
3DCrowdNet, designed to address the domain gap by leveraging a crowded scene-robust image feature, is commonly employed for its robust 3D pose estimation performance in in-the-wild crowded scenes.
4DHumans, which achieves leading 3D human mesh recovery through a transformer architecture, is distinguished by its exceptional performance on challenging poses.
To ensure the broad applicability of our approach, we conduct experiments across these diverse TPEs, which represent different architectures and target scenarios, reflecting the varying model sizes and constraints encountered in real-world applications.

We compared our method against PoseGen~\cite{gholami2024posegen} and \textit{Ours-N}, two baselines focused on augmenting 3D pose data. 
Unlike \textit{Ours}, which measures TPE performance on each real-world image to identify challenging poses where TPE underperforms, PoseGen generates synthetic poses with a GAN and renders them via NeRF, identifying challenging poses by measuring TPE performance on these artificially created images.
\textit{Ours-N} employs the same challenging-pose identification strategy (EEM) and motion synthesis (MSM) as \textit{Ours}, but replaces the human animation model~\cite{zhu2024champ} with NeRF rendering--mirroring PoseGen's image-generation method.
Thus, comparing \textit{Ours} and \textit{Ours-N} reveals the importance of realistic image rendering, while comparing \textit{Ours-N} and PoseGen highlights the advantage of our key modules (\ie, EEM and MSM).

\subsection{Datasets}
Before applying our approach, we pre-trained each TPE (\ie, Hybrik, 3DCrowdNet, and 4DHumans) on five datasets: Human3.6M~\cite{h36m_pami}, MuCo~\cite{mehta2018single}, MPI-INF-3DHP~\cite{mehta2017monocular}, MPII~\cite{andriluka14cvpr}, and MSCOCO~\cite{lin2014microsoft}.
The first three datasets (\ie, Human3.6M, MuCo, and MPI-INF-3DHP) provide accurate 3D pose annotations captured with multi-view camera systems, but they either lack background diversity or have synthetic backgrounds.
On the other hand, MPII and MSCOCO offer more diverse and complex backgrounds, but only contain pseudo-labeled~\cite{Moon_2023_CVPRW_3Dpseudpgts} 3D pose annotations. 
After pre-training TPE on these datasets, our framework synthesized an additional 3D pose dataset from the MPII dataset, an in-the-wild 2D pose dataset.
Specifically, we identified the top 500 challenging samples (${K}_{\text{C}}=500$) and the bottom 200 non-challenging samples (${K}_{\text{NC}}=200$) from single-person in-the-wild 2D pose data in the MPII dataset.
Then, using our proposed methodology including the filtering based on Eq.~\ref{eq:filtering_error}, we synthesized 27,000 3D pose data samples from the source 2D data.
For evaluation, we utilized six datasets: 3DPW~\cite{vonMarcard2018}, EMDB~\cite{kaufmann2023emdb}, CMU~\cite{Joo_2017_TPAMI}, HuMMan~\cite{cai2022humman}, LSPET~\cite{johnson2011learning, johnson2010clustered}, and JHMDB~\cite{Jhuang:ICCV:2013}.
These datasets encompass various human appearances and poses in in-the-wild scenarios~\cite{vonMarcard2018,kaufmann2023emdb,johnson2011learning,johnson2010clustered,Jhuang:ICCV:2013} or multi-view camera setups~\cite{Joo_2017_TPAMI,cai2022humman}.
For 3D pose evaluation on the 3DPW, EMDB, CMU, and HuMMan datasets, we used two metrics: Mean Per Joint Position Error (MPJPE~\cite{h36m_pami}) and Procrustes-aligned MPJPE (PA-MPJPE~\cite{h36m_pami}).
For 2D pose evaluation on the LSPET and JHMDB datasets, we used PCKh~\cite{andriluka14cvpr} metric.

\subsection{Comparison with Baselines}

\begin{figure}[tb]
\begin{center}
\centerline{\includegraphics[width=0.9\columnwidth]
{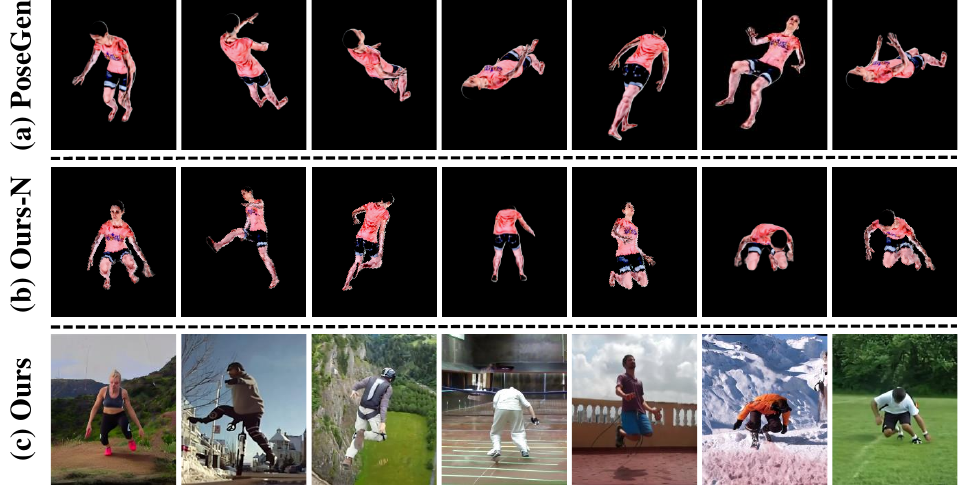}}
\vskip -0.1in
\caption{\textbf{Qualitative Comparison of Augmented Data.} 
(a) \textbf{PoseGen} generates 3D pose data where poses differ from real-world problematic poses due to the way of acquiring challenging poses from synthetic images rendered by NeRF.
(b) \textbf{Ours-N} extracts challenging poses from in-the-wild dataset but lacks diversity in rendered images.
(c) \textbf{Ours} not only acquires problematic poses from real-world but also generates images with various human appearances and realistic backgrounds. Note that each column in (b) and (c) displays images generated from the same poses.
}
\label{fig:data_view}
\end{center}
\vskip -0.35in
\end{figure}

We evaluated PoseSyn through both qualitative and quantitative comparisons against two baselines across three TPEs. 
As shown in Fig.~\ref{fig:main_qual_image}, our methodology, along with PoseGen and \textit{Ours-N}, improved the performance of TPE (\ie, 3DCrowdNet).
However, PoseGen underperformed both \textit{Ours-N} and \textit{Ours} on the sample from the JHMDB dataset in the first row of Fig.~\ref{fig:main_qual_image}.
This discrepancy occurs because PoseGen obtains challenging poses from NeRF-rendered images, as shown in Fig.~\ref{fig:data_view}\,(a), rather than directly from in-the-wild images.
Furthermore, even when challenging poses were obtained from in-the-wild images, \textit{Ours} surpassed \textit{Ours-N} on the sample from the EMDB dataset in the second row.
This improvement stems from the fact that the 3D pose data generated by \textit{Ours-N} primarily consists of NeRF-rendered images, limiting diversity in both backgrounds and human appearances--an important element for training image-to-3D pose estimators--as illustrated in Fig.~\ref{fig:data_view}\,(b).
In contrast, as depicted in Fig.~\ref{fig:data_view}\,(c), PoseSyn not only acquires challenging poses from real-world data, but also leverages a human animation model to generate images that showcase various human appearances and intricate poses with realistic backgrounds.
Consequently, PoseSyn demonstrated significantly improved performance, as illustrated in Fig.~\ref{fig:main_qual_image}, aligning with quantitative metrics in Tab.~\ref{table1}.
Our method yields 6--14\% improvements in MPJPE and 5--9\% in PA-MPJPE on 3D datasets, along with a 1--2\% accuracy gain on 2D datasets across various TPEs.
These results confirm that PoseSyn provides robust data augmentation benefits, consistently enhancing performance across various real-world scenarios.

\subsection{Effectiveness of MSM}
\label{sec:effect_pam}

\textbf{\textit{
Does MSM effectively augment identified challenging poses into motion sequences?
}}
To evaluate whether MSM effectively augments challenging poses identified by EEM into motion sequences, we conducted a direct evaluation of how closely MSM-generated motions align with the GT 3D poses--\textit{without} training the TPE on these synthesized data.
We first selected challenging samples from EMDB dataset, which contains GT 3D pose annotations, by applying TPE (3DCrowdNet) and choosing the top 100 samples with the largest error.
On these challenging samples, the average PA-MPJPE between the TPE-predicted and GT poses was 181.7 mm (Tab.~\ref{Tab:MSM_effectiveness}\,(a)).
We then augmented each of 100 TPE-predicted poses into motion sequences using two methods: (b) MSM without initial motion representation (\ie, w/o $\mathcal{MR}_{\text{init}}$) and (c) our full MSM with both initial motion and text description.
Within each sequence, we computed the PA-MPJPE between every synthesized pose and the GT pose, then calculated the mean, standard deviation (std), and minimum (min) of these PA-MPJPE values. 
Finally, we averaged these three statistics across all 100 sequences.
Since (a) is just a single TPE-predicted pose for each hard case, its mean and min are the same and its std is zero.
As shown in Tab.~\ref{Tab:MSM_effectiveness}, though motion synthesis approaches (\ie, (b) and (c)) have higher mean error than mis-prediction (a), these approaches achieved lower min error. This indicates that, by synthesizing plausible pose variants, they produce at least one pose that is closer to the true problematic pose than the single naive mis-predicted pose.
Notably, (b) omitting the initial pose representation raises both mean (222.3 mm) and min (151.1 mm) error, emphasizing the vital role of $\mathcal{MR}_{\text{init}}$ to accurately target challenging poses.

\begin{table}[t]
\centering
\resizebox{0.9\columnwidth}{!}{
\begin{tabular}{l|l|l}
\hline
          \footnotesize{\textbf{Method}} & \footnotesize{\textbf{Mean}\,($\downarrow$) \, $\pm$ \, \textbf{Std} ($\uparrow$)} & \footnotesize{\textbf{Min} ($\downarrow$)}                 \\ \hline \hline
         
\footnotesize{(a)} \textbf{$\hat{J}^{\text{3D}}$} & \footnotesize{\textbf{181.7} $\pm$ 0.0 mm} & \footnotesize{181.7 mm} \\ \hline 
\footnotesize{(b)} \footnotesize{\textbf{w/o $\mathcal{MR}_{\text{init}}$}} & \footnotesize{222.3 $\pm$ 36.4 mm} & \footnotesize{151.1 (-16.8\%) mm} \\ \rowcolor[HTML]{EFEFEF}
\footnotesize{(c)} \textbf{Ours} & \footnotesize{209.3 $\pm$ \textbf{36.5} mm} & \footnotesize{\textbf{140.8 (-22.5\%)} mm} \\ \hline
\end{tabular}
}
\vskip -0.10in
\caption{\textbf{Analysis on Effectiveness of MSM.}
We assessed how well the generated poses reflect the identified challenging poses. This evaluation was conducted using PA-MPJPE between the GT 3D pose in EMDB dataset and the generated poses. Results show (a) mis-prediction, (b) without $\mathcal{MR}_{\text{init}}$, (c) our full MSM.
}
\vskip -0.15in
\label{Tab:MSM_effectiveness}
\end{table}

\begin{table}[t]
\centering
\resizebox{1\linewidth}{!}{
\begin{tabular}{c|l|ccccc|c}
\hline
\multirow{2}{*}{\textbf{Method}}& \multirow{2}{*}{\textbf{Metrics}} & \multirow{2}{*}{\textbf{3DPW}} & \multirow{2}{*}{\textbf{EMDB}} & \textbf{CMU} & \textbf{CMU} & \multirow{2}{*}{\textbf{HuMMan}} & \multirow{2}{*}{\textbf{Mean}}\\
& &                     &                     & \textbf{171204} & \textbf{171206} & &
\\ \hline\hline
 & \footnotesize {MPJPE} $\downarrow$ & 81.7 & 115.8 & 108.8 & 110.7 & 98.9 & 103.2\\
\multirow{-2}{*}{ Real-only}                            & \footnotesize {PA-MPJPE} $\downarrow$ &51.1 & 71.2  & 72.5 & 70.4 & 65.8 & 66.2\\\hline
\multirow{2}{*}{$\hat{{J}}^{\text{3D}}$} & \footnotesize {MPJPE} $\downarrow$ & 78.7 & 115.8   & 102..7 & $\textbf{104.8}$ & 96.2 & 99.6\\
                                                 & \footnotesize {PA-MPJPE} $\downarrow$ &51.1 & 72.1  & 68.4 & 69.6 & 65.6 & 65.4\\
                                                  & \footnotesize {MPJPE} $\downarrow$ & 78.2 & 112.7 & 102.6 & 106.1 & \textbf{93.1}  & 98.5\\
\multirow{-2}{*}{w/o $\mathcal{MR}_{\text{init}}$} & \footnotesize {PA-MPJPE} $\downarrow$ &50.1 & 70.2  & 68.6 & 68.9 & 64.0         & 64.4\\
\rowcolor[HTML]{EFEFEF} & \footnotesize {MPJPE} $\downarrow$ & $\textbf{77.4}$ & $\textbf{111.0}$ & $\textbf{101.0}$ & 105.0 & $\textbf{93.1}$ & $\textbf{97.5}$\\ 
\rowcolor[HTML]{EFEFEF} \multirow{-2}{*}{Ours} & \footnotesize {PA-MPJPE} $\downarrow$ & $\textbf{48.9}$ & $\textbf{68.3}$ & $\textbf{67.3}$ & $\textbf{67.9}$ & $\textbf{62.3}$ & $\textbf{62.9}$\\\hline
\end{tabular}
}
\vskip -0.1in
\caption{\textbf{Effectiveness of MSM in TPE Training.}
We assessed the impact of MSM on TPE performance by training 3DCrowdNet using data augmented with three different MSM configurations (mis-prediction, without $\mathcal{MR}_{\text{init}}$, and our full MSM).
}
\vskip -0.25in
\label{tab:pam}
\end{table}

\begin{figure}[t]
\begin{center}
\centerline{\includegraphics[width=0.9\linewidth]{
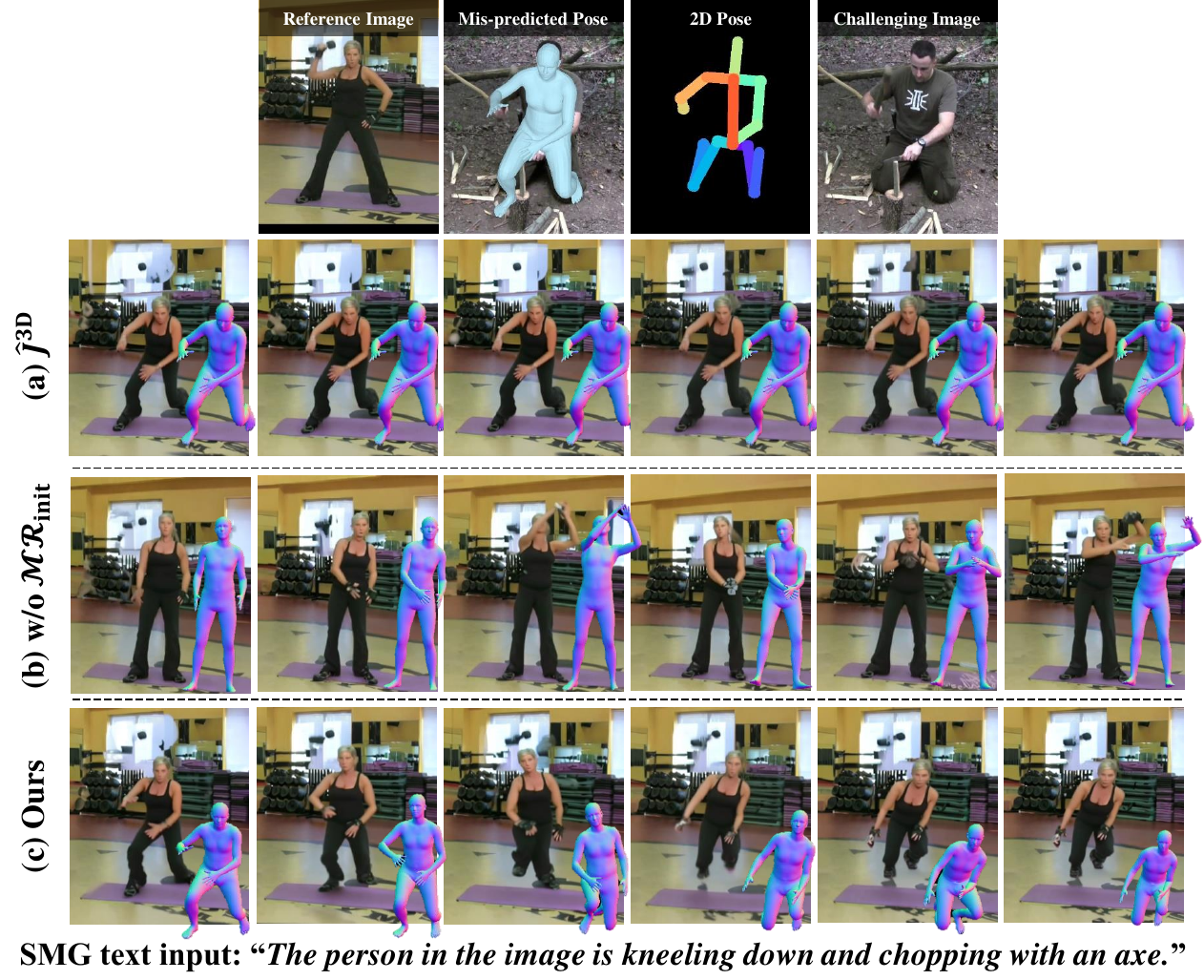
}
}
\vskip -0.15in
\caption{
\textbf{Effectiveness of MSM in Data Synthesis.} 
The first row shows the challenging image and pose in $\mathcal{D}_{\text{C}}$ and reference image in $\mathcal{D}_{\text{NC}}$. (a) Without MSM, images are generated using the repeated mis-predicted pose as a condition. (b) Without $\mathcal{MR}_{\text{init}}$, MSM-generated motion only reflects ``chopping with an axe'' missing ``kneeling down'', the actual problematic pose. (c) Ours effectively augments the actual problematic pose into 3D pose data.
}
\label{fig:MSM_multi}
\vskip -0.35in
\end{center}
\end{figure}

\textbf{\textit{
Is the effectiveness of MSM essential for TPE's performance improvement?
}}
MSM aims to augment the mis-predicted pose for challenging image into motion sequences better approximate the actual challenging pose, followed by human animation model to generate images under condition of the augmented motion sequences.
Without MSM, images are generated using only the mis-predicted pose as a condition, as shown in Fig.~\ref{fig:MSM_multi}\,(a).
Since this mis-predicted pose differs from the actual challenging pose, 3D pose data synthesized in this manner does not properly reflect such challenging scenarios. 
Furthermore, this approach has no ability to produce any diverse pose variations for these difficult cases, leading to minimal performance gains when fine-tuning with such monotonous pose data, as indicated in Tab.~\ref{tab:pam} (\ie., $\hat{{J}}^{\text{3D}}$).
Conversely, when images are generated using augmented motion that aligns more closely with the challenging images--thanks to MSM--the performance improvement was maximized (See Ours in Tab.~\ref{tab:pam}). 
This is because the human animation model generates images conditioned on motions close to pose where TPE shows low performance, effectively targeting those challenging scenarios.
Lastly, without initial motion representation, generated motions failed to reflect the actual problematic pose's configuration as shown in Fig.~\ref{fig:MSM_multi}\,(b), leading to limited performance gains (\ie, w/o $\mathcal{MR}_{\text{init}}$ in Tab.~\ref{tab:pam}).

\subsection{Ablation Study}
\label{sec:ablation}

\begin{table}[t]
\centering
\resizebox{1\linewidth}{!}{
\begin{tabular}{c|c|l|ccccc|c}
\hline
\textbf{Model} & \textbf{Training}  & \multirow{2}{*}{\textbf{Metrics}} & \multirow{2}{*}{\textbf{3DPW}} & \multirow{2}{*}{\textbf{EMDB}} & \textbf{CMU} & \textbf{CMU} & \multirow{2}{*}{\textbf{HuMMan}} & \multirow{2}{*}{\textbf{Mean}}\\
\textbf{(EEM)} & \textbf{method} & &                     &                     & \textbf{171204} & \textbf{171206} & & \\
\hline\hline
 & & \footnotesize{MPJPE} $\downarrow$ & 80.2 & 114.2 & 107.1 & 109.0 & 94.9 & 101.1\\
\multirow{-2}{*}{w/o EEM} & \multirow{-2}{*}{FT} & \footnotesize{PA-MPJPE} $\downarrow$ & 50.5 & 71.1 & 72.0 & 71.1 & 65.1 & 66.0\\
\hline
\multirow{2}{*}{Hybrik} & \multirow{2}{*}{FT} & \footnotesize{MPJPE} $\downarrow$ & 79.3 & 113.5 & 103.5 & 106.2 & 94.9 & 99.5\\
 & & \footnotesize{PA-MPJPE} $\downarrow$ & 50.0 & 70.4 & 69.4 & 68.8 & 64.6 & 64.6\\
\multirow{2}{*}{4DHumans} & \multirow{2}{*}{FT} & \footnotesize{MPJPE} $\downarrow$ & 78.6 & 112.7 & 103.0 & 106.6 & 93.2 & 98.8\\
 & & \footnotesize{PA-MPJPE} $\downarrow$ & 50.0 & 69.9 & 68.4 & 68.4 & 64.2 & 64.2\\
\multirow{2}{*}{3DCrowdNet} & \multirow{2}{*}{FS} & \footnotesize{MPJPE} $\downarrow$ & 78.8 & 112.4 & 103.0 & 106.2 & 93.5 & 98.8\\
 & & \footnotesize{PA-MPJPE} $\downarrow$ & 49.5 & 69.9 & 68.2 & 67.9 & 63.7 & 63.8\\
\rowcolor[HTML]{EFEFEF} & & \footnotesize{MPJPE} $\downarrow$ & \textbf{77.4} & \textbf{111.0} & \textbf{101.0} & \textbf{105.0} & \textbf{93.1} & \textbf{97.5}\\
\rowcolor[HTML]{EFEFEF}\multirow{-2}{*}{3DCrowdNet} & \multirow{-2}{*}{FT} & \footnotesize{PA-MPJPE} $\downarrow$ & \textbf{48.9} & \textbf{68.3} & \textbf{67.3} & \textbf{67.9} & \textbf{62.3} & \textbf{62.9}\\
\hline
\end{tabular}

}
\vskip -0.10in
\caption{\textbf{Ablation Study on EEM.}
We conducted an ablation study to examine the effect of different pose estimators within EEM on TPE performance. 
We compared ours with three settings: (1) excluding EEM entirely, (2) using a pose estimator different from the TPE (\ie, Hybrik and 4DHumans) in EEM, and (3) using the same TPE model, 3DCrowdNet, in EEM but training estimator from scratch. ``FT" denotes a fine-tuning approach and ``FS" represents an approach of training estimator from scratch. 
}
\vskip -0.25in
\label{tab:ablation_study_EEM}
\end{table}

To assess the effectiveness of EEM, we first ablated EEM in our framework.
As shown in Tab.~\ref{tab:ablation_study_EEM}, without EEM, the performance gain was minimal, highlighting the importance of identifying and synthesizing challenging samples where TPE underperforms.
Additionally, we evaluated how using a \textit{different} TPE inside EEM would affect final performance on the actual TPE (\ie, 3DCrowdNet).
Specifically, we replaced 3DCrowdNet with Hybrik or 4DHumans for identifying challenging poses.
Although both variants showed improved performance upon the without EEM baseline, neither matched the performance achieved by using 3DCrowdNet in EEM, which is our original method.
This demonstrates that each TPE has its own problematic poses with lower performance, so focusing data synthesis on these poses yields better results.
We also tested an approach called, 3DCrowdNet~(FS), which uses a pre-trained 3DCrowdNet to identify hard samples, but trains a randomly initialized 3DCrowdNet using synthesized data from those hard samples.
While 3DCrowdNet~(FS) surpassed different model architectures, it still underperformed compared to using the exact same TPE weights.
Overall, these results confirm that using the same pre-trained TPE within EEM is the most effective way to generate tailored training data and maximize generalization gains.
Lastly, we conducted an ablation study for various threshold values in the filtering step, allowing us to assess its impact on training the TPE with our created 3D pose data, as shown in Tab~\ref{tab:ablation_filtering}. 
Though our method utilizes an off-the-shelf human animation model for creating images aligned with synthesized motion, editing by motion can sometimes lead to blending artifacts between the human and background. 
Using the degraded data blindly (high $\tau$) can lower model performance, while excessive filtering (low $\tau$) could leave only simpler samples, restricting performance improvement.
We set $\tau=120$ to ensure high-quality images for training the image-to-3D pose estimator, while still retaining problematic pose samples where TPE struggles.

\begin{table}[t]
\centering
\resizebox{1\columnwidth}{!}{
\begin{tabular}{c|c|cccc|cc}
\hline
\multicolumn{2}{c|}{\textbf{Filtering}}
                   & \multicolumn{2}{c|}{\textbf{3DPW}}                                                         & \multicolumn{2}{c|}{\textbf{EMDB}}          & \multicolumn{2}{c}{\textbf{Mean}}                            \\ \hline
\multicolumn{1}{c|}{$\boldsymbol{\tau}$} & 
\multicolumn{1}{c|}{\textbf{Pass}} & {\small\textbf{MPJPE}$\downarrow$} & \multicolumn{1}{c|}{\small\textbf{PA-MPJPE}$\downarrow$} & \multicolumn{1}{c|}{\small\textbf{MPJPE}$\downarrow$} & \multicolumn{1}{c|}{\small\textbf{PA-MPJPE}$\downarrow$} & \multicolumn{1}{c|}{\small\textbf{MPJPE}$\downarrow$} & \small\textbf{PA-MPJPE$\downarrow$} \\ \hline\hline

\rowcolor[HTML]{EFEFEF} 
200 & 72.6\%              & 79.9                                & 50.2                                   & 114.5                               & 70.8         & 97.2 &  61.0                        \\
160 & 64.0\%               & 78.7                                & 49.7                                   & 113.0                               & 70.3         & 95.9 &  60.0                        \\
\rowcolor[HTML]{EFEFEF} 
120 & 45.7\%       & \textbf{77.4}                       & \textbf{48.9}                          & \textbf{111.0}                      & \textbf{68.3} & \textbf{94.2} & \textbf{58.6} \\
80 & 28.2\%                & 78.6                                & 49.6                                   & 111.9                               & 69.8         & 95.3 & 59.7                          \\
\rowcolor[HTML]{EFEFEF} 
40 & 6.6\%                 & 79.5                                & 49.8                                   & 112.4                               & 70.1          & 96.0 &  60.0                       \\
\hline

\end{tabular}
}
\vskip -0.10in
\caption{\textbf{Ablation Study on Filtering.}
We experimented with different threshold values ($\tau$) in filtering process to examine their impact on TPE (\ie 3DCrowdNet) performance when training with our synthesized 3D pose data.
``Pass" denotes the percentage of data retained after filtering step.
}
\vskip -0.2in
\label{tab:ablation_filtering}
\end{table}

%% file: folder_tex/Conclusion_arxiv.tex
We present PoseSyn, a novel framework to improve TPE generalization by effectively augmenting in-the-wild 2D pose data into enriched 3D pose training data. 
Our EEM effectively identifies challenging poses and images, while our MSM generates motion sequences aligned with the identified problematic poses under the guidance of both text description and initial motion representation. 
These motion sequences are then used to generate images of various human appearances and poses, enriching 3D pose data and enhancing the generalization capability of TPEs.
Experimental results confirm that PoseSyn achieves notable accuracy improvements across various benchmarks and TPE architectures. 
We believe this framework provides a promising solution for enhancing generalization of pose estimator regardless of its model characteristic without the need for costly acquisition of 3D pose data.
\vspace{-3pt}
\paragraph{Limitation}
Our framework utilizes an off-the-shelf human animation model~\cite{zhu2024champ} for motion-guided video generation, editing a human in a reference image based on generated motions.
However, since the animation model struggles with multi-person scenarios (\eg, human interaction), our framework is also limited to single-person scenario.
Extending our framework would require advancements in multi-person animation model with the integration of interaction-aware motion generation models into our MSM.
Future work could focus on these developments to support multi-person 3D pose data synthesis.

%% file: supplementary_main_arixv.tex
% \input{preamble}
% \begin{document}
% \maketitle

% \appendix

\section{Implementation Details}
\label{implementation_details}

\paragraph{Implementation Details}
We selected three image-to-3D pose estimation models as our target pose estimators (TPEs): Hybrik~\cite{li2021hybrik}, 3DCrowdNet~\cite{choi2022learning}, and 4DHumans~\cite{goel2023humans}.
All experiments were conducted utilizing each official code of 3DCrowdNet\footnote{\url{https://github.com/hongsukchoi/3DCrowdNet_RELEASE}}, Hybrik\footnote{\url{https://github.com/Jeff-sjtu/HybrIK}} and 4DHumans\footnote{\url{https://github.com/shubham-goel/4D-Humans}}.

Specifically, for 3DCrowdNet, we downloaded a pre-trained model which was trained for 10 epochs on real datasets (\eg, Human36M~\cite{h36m_pami}, MuCo~\cite{mehta2018single}, MSCOCO~\cite{lin2014microsoft}, and MPII~\cite{andriluka14cvpr} datasets) with a learning rate of $1\times10^{-4}$.
We then applied our data synthesis framework to augment the MPII dataset, creating a 3D pose dataset with 27,291 samples.
We fine-tuned the 3DCrowdNet on this synthesized dataset with a batch size of 64 and a learning rate of $1\times10^{-5}$ for 10 epochs, utilizing both real data and synthesized data.

In the case of 4DHumans model,  we downloaded a pre-trained model which was trained for 1M iterations on real datasets (\eg, Human36M, MPI-INF-3DHP~\cite{mehta2017monocular}, AVA~\cite{gu2018ava}, AIC~\cite{wu2017ai}, INSTA~\cite{kanazawa2019learning}, MSCOCO, and MPII datasets) with a learning rate of $1\times10^{-4}$.
We then applied our data synthesis framework to augment the MPII dataset, creating a 3D pose dataset with 27,872 samples.
We fine-tuned the 4DHumans on this created dataset with a batch size of 32 and a learning rate of $1\times10^{-5}$ for 200K iterations, utilizing both real data and created data.

Likewise for Hybrik, we downloaded a pre-trained model which was trained for 120 epochs on real datasets (\eg, Human36M, MPI-INF-3DHP, and MSCOCO datasets) with a learning rate of $1\times10^{-3}$. We then applied our data synthesis framework to obtain a synthesized 3D pose dataset consisting of 26,758 samples, and fine-tuned the Hybrik with a batch size of 64 and a learning rate of $1\times10^{-4}$ for 40 epochs utilizing both real data and synthesized data.
Note that each synthesized 3D dataset differs per model because challenging images are extracted from each TPE.

Additional parameters, including the number of challenging data (\ie, $N_{\text{C}}$), the number of non-challenging data (\ie, $N_{\text{NC}}$), and the filtering threshold $\tau$, are provided in Tab.~\ref{tab:notation}.

\begin{figure}[ht]
\vskip 0.2in
\begin{center}
\centerline{\includegraphics[width=\linewidth]
{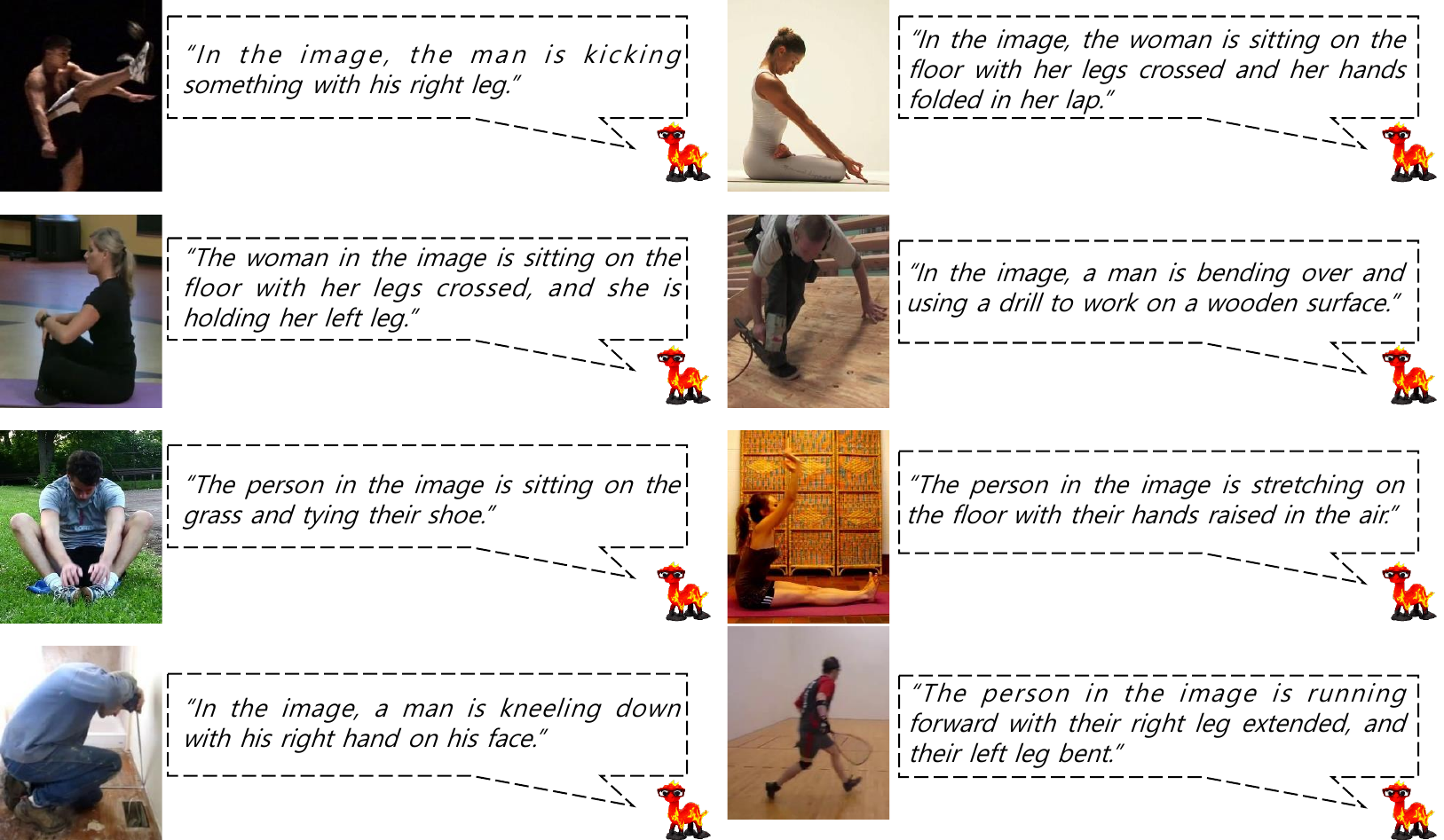}}
\vskip -0.05in
\caption{\textbf{VLM Prompting.}
We visualize challenging image and extracted textual description pairs.
}
\label{fig:vlm_answering}
\end{center}
\vskip -0.2in
\end{figure}

\paragraph{VLM Prompting}
In our proposed Semantic-guided Motion Generation (SMG) stage, we augments an identified challenging pose into motion sequences guided by both textual description and initial pose representation. 
To extract the textual description of the challenging image, we leverage a VLM~\cite{wang2024llavallama3} to ask the question, \textit{``What is the motion of the someone in the image? Please answer similar to \{Text-to-Motion prompt examples\}"}, where \{\textit{Text-to-Motion prompt examples}\} are provided as follows:
\begin{itemize}
\item ``A man kicks something or someone with his left leg."
\item ``A person walking forward and then turns around."
\item ``A person squats down then jumps."
\item ``A person raises their right hand to their face."
\item ``He is waving with his right hand."
\item ``A person kicks something with their right foot."
\item ``A running man hops over something a comes down to a walk."
\item ``A man raises both arms, then kneels down."
\item ``Person is using their left arm to dodge a punch."
\item ``A person raised both their arms and started to clap."
\end{itemize}
We visualize the results of VLM answers in Fig.~\ref{fig:vlm_answering}.
\paragraph{Details for Orientation-aligned Motion Guidance}
PoseSyn synthesizes human animated images $\mathcal{V}_{\text{C}}=\{I_{\text{C}, l}\}_{l=1}^{L}$ by using non-challenging image $I_\text{NC}$ as reference image and generated motion sequences $\mathcal{M}_{\text{C}}=\{{J}^{\text{3D}}_{\text{C}, l}\}_{l=1}^{L}$ as motion guidance.
To ensure natural human animation, we align the global orientation of $\mathcal{M}_{\text{C}}$ with the human orientation in ${I}_{\text{NC}}$.
First, following Champ~\cite{zhu2024champ}'s original method, we exploit 4DHumans to obtain camera parameters (\ie, focal length $f$ and principal points $p$) and SMPL parameters including global orientation $\theta^{\text{G}} \in \mathbb{R}^{3}$, pose $\theta^{\text{P}} \in \mathbb{R}^{69}$, shape $\beta \in \mathbb{R}^{10}$, and translation $\bar{t} \in \mathbb{R}^{3}$ for the reference image $I_\text{NC}$, which is denoted as:
\begin{equation}
\theta^{\text{G}},\theta^{\text{P}},\beta,\bar{t},f,p=\text{4DHumans}({I}_{\text{NC}}).
\end{equation}
As Champ utilizes rendered meshes of the motion sequences as motion guidance, we obtain SMPL parameters $\{{\theta^{\text{G}}_{\text{S}, l} } ,{\theta^{\text{P}}_{\text{S}, l}}\}^{L}_{l=1}$ from the 3D joint sequences $\mathcal{M}_{\text{C}}$ by applying an optimization method~\cite{zhang2023generating} $O$ as follows: 
\begin{equation}
\{\theta^{\text{G}}_{\text{S}, l} ,\theta^{\text{P}}_{\text{S}, l}\}^{L}_{l=1}=O(\mathcal{M}_{\text{C}}),
\end{equation}
where SMPL parameters are represented in the Rodrigues' rotation form~\cite{loper2023smpl}.
To align the global orientation of $\mathcal{M}_{\text{C}}$ with the human orientation in ${I}_{\text{NC}}$, we use the global orientation $\theta^{G}$ of $\mathcal{M}_{\text{C}}$ as the initial global orientation for the first motion sequence.
For the remaining frames, we adjust the initial global orientation based on the relative angle between the global orientations of the motion sequences.
Specifically, let's denote $T_{r\rightarrow e}(\cdot)$ as the conversion from the Rodrigues' rotation form to Euclidean one, $R_{\text{S},l}=T_{r\rightarrow e}(\theta^{\text{G}}_{\text{S},l})$ as the rotation matrix for the global orientation represented in Euclidean space, and then the modified global orientation sequences are formulated as follows:

\begin{equation}
    \quad \theta^{\text{G}}_{\text{mod}, l}=T_{r\rightarrow e}^{-1}(R_{\text{S}, l}(R_{\text{S}, l-1})^{-1}R_{\text{init}}), \quad l=2,...,L,
\end{equation}
where $\theta^{\text{G}}_{\text{mod}, 1}=\theta^{\text{G}}$ is the initial global orientation and $R_{\text{init}}=T_{r\rightarrow e}(\theta^{\text{G}})$ is its representation in Euclidean space. 
Then, we acquire orientation-aligned mesh sequences $\{Mesh_{l}\}^{L}_{l=1}$, with each orientation-aligned mesh defined as:
\begin{equation}
    Mesh_{l}=SMPL(\theta^{\text{G}}_{\text{mod}, l}, \theta^{\text{P}}_{\text{S}, l},\beta,\bar{t}).
\end{equation}
Finally, we project each mesh onto the image plane by using the camera parameters for the reference image $I_{\text{NC}}$ to obtain a sequence of orientation-aligned rendered mesh $\{I_{\text{render}, l}\}^{L}_{l=1}$, where each rendered mesh is calculated as follows:  
\begin{equation}
    I_{\text{render}, l}=\text{Proj}(Mesh_{l},f,p).
\end{equation}
This rendered mesh sequence serves as motion guidance for animating a human in the reference image $I_\text{NC}$. For more details, please refer to \cite{zhu2024champ}.

\clearpage
\newpage

\renewcommand{\arraystretch}{1.2}
\input{folder_tex/Notation_Table}
\renewcommand{\arraystretch}{1.0}

\clearpage
\newpage

\vskip 0.3in
\section{Extendibility of PoseSyn}

To ensure a fair comparison with the baselines, we utilized the MPII dataset~\cite{andriluka14cvpr}, which was originally used to pre-train the TPE (3DCrowdNet), as the basis for synthesizing the 3D pose dataset throughout our experiments. 
However, our methodology is not restricted to the data used for pre-training the TPE; rather, it can augment any easily accessible in-the-wild unlabeled images into a 3D pose dataset, showcasing its flexibility and broad applicability.

To validate this capability, we assigned pseudo-label annotations to each image in the dataset using either a 2D pose estimator~\cite{cao2017realtime} or a 3D pose estimator~\cite{cha2022multi}, rather than relying on the 2D GT poses from the MPII dataset. 
In the case of 2D pseudo-labeling, the EEM module calculated the error as outlined in Eq.~(1), which was then used to identify challenging and non-challenging datasets as expressed in Eq.~(2) and Eq.~(3), respectively.

For the 3D pseudo-labeling, the criterion metric $Err$ was computed by replacing 2D joints with their 3D counterparts in Eq.~(1), followed by the same methodology for identifying challenging and non-challenging samples.

\begin{table}[b]
\centering
\resizebox{1\columnwidth}{!}{
\begin{tabular}{ccccccccccc|cc}
\hline
\rowcolor[HTML]{EFEFEF} 
\multicolumn{1}{c|}{\cellcolor[HTML]{EFEFEF}}                      & \multicolumn{2}{c|}{\cellcolor[HTML]{EFEFEF}\textbf{3DPW}}                                                                                                                                 & \multicolumn{2}{c|}{\cellcolor[HTML]{EFEFEF}\textbf{EMDB}}
          & \multicolumn{2}{c|}{\cellcolor[HTML]{EFEFEF}\textbf{CMU\_171204}}                                                             & \multicolumn{2}{c|}{\cellcolor[HTML]{EFEFEF}\textbf{CMU\_171026}}                                                             & \multicolumn{2}{c|}{\cellcolor[HTML]{EFEFEF}\textbf{HuMMan}} & \multicolumn{2}{c}{\cellcolor[HTML]{EFEFEF}\textbf{Mean}} \\ \hline
\rowcolor[HTML]{FFFFFF} 
\multicolumn{1}{c|}{\cellcolor[HTML]{FFFFFF}\textbf{}}             & \multicolumn{1}{c|}{\cellcolor[HTML]{FFFFFF}\footnotesize\textbf{MPJPE$\downarrow$}} & \multicolumn{1}{c|}{\cellcolor[HTML]{FFFFFF}\footnotesize\textbf{PA-MPJPE$\downarrow$}} & \multicolumn{1}{c|}{\cellcolor[HTML]{FFFFFF}\footnotesize\textbf{MPJPE$\downarrow$}}    & \multicolumn{1}{c|}{\cellcolor[HTML]{FFFFFF}\footnotesize\textbf{PA-MPJPE$\downarrow$}}   & \multicolumn{1}{c|}{\cellcolor[HTML]{FFFFFF}\footnotesize\textbf{MPJPE$\downarrow$}} & \multicolumn{1}{c|}{\cellcolor[HTML]{FFFFFF}\footnotesize\textbf{PA-MPJPE$\downarrow$}} & \multicolumn{1}{c|}{\cellcolor[HTML]{FFFFFF}\footnotesize\textbf{MPJPE$\downarrow$}}    & \multicolumn{1}{c|}{\cellcolor[HTML]{FFFFFF}\footnotesize\textbf{PA-MPJPE$\downarrow$}} & \multicolumn{1}{c|}{\cellcolor[HTML]{FFFFFF}\footnotesize\textbf{MPJPE$\downarrow$}}    & \multicolumn{1}{c|}{\cellcolor[HTML]{FFFFFF}\footnotesize\textbf{PA-MPJPE$\downarrow$}}& \multicolumn{1}{c|}{\cellcolor[HTML]{FFFFFF}\footnotesize\textbf{MPJPE$\downarrow$}}    & \multicolumn{1}{c}{\cellcolor[HTML]{FFFFFF}\footnotesize\textbf{PA-MPJPE$\downarrow$}}  \\ \hline
\rowcolor[HTML]{EFEFEF} 
\multicolumn{1}{c|}{\cellcolor[HTML]{EFEFEF}Real-only}     & 81.7                                                        & 51.1                                                                                                         & 115.8                                                          & 71.2      &108.8 & 72.5 &110.7 &70.4 & 98.9& 65.8 &103.2 &66.2                                                                   \\
\rowcolor[HTML]{FFFFFF} 
\multicolumn{1}{c|}{\cellcolor[HTML]{FFFFFF}3D Pose Estimator}     & 78.9                                                        & 49.8                                                                                                        & 112.5                                                          & 69.9 &103.6 &69.5 &106.6 &69.9& 96.5 &63.9&99.6 &64.6                                                                    \\
\rowcolor[HTML]{EFEFEF} 
\multicolumn{1}{c|}{\cellcolor[HTML]{EFEFEF}2D Pose Estimator}     & 78.5                                                        & 49.7                                                                                                    & 112.2                                                          & 69.9      &103.3 &69.6 &106.4 &68.5 &\textbf{93.1} &63.7&98.7 &64.3                                                        \\
\rowcolor[HTML]{FFFFFF} 
\multicolumn{1}{c|}{\cellcolor[HTML]{FFFFFF}\textbf{Ground Truth}} & \textbf{77.4}                                               & \textbf{48.9}                                                                                          & \textbf{111.0}                                                 & \textbf{68.3}                                              &\textbf{101.0} &\textbf{67.3}& \textbf{105.0} &\textbf{67.9} &\textbf{93.1} &\textbf{62.3} &\textbf{97.5} &\textbf{62.9}\\ \hline

\end{tabular}
}
\caption{\textbf{EEM Approach Variants.}
We report results in the performance improvement of TPE when trained with synthesized dataset obtained through different EEM approaches: 3D pose pseudo-labeling, 2D pose pseudo-labeling, and ground truth (GT) annotations, where GT annotations represent the original PoseSyn method.
The Real-only model, trained exclusively on real datasets, is used as the reference model and fine-tuned with each synthesized dataset.
MPJPE and PA-MPJPE metrics are reported across multiple datasets to demonstrate the impact of the synthesis methods on TPE performance.
}
\label{tab:pseudo_label}
\end{table}

\clearpage
\newpage

As shown in Tab.~\ref{tab:pseudo_label}, our method demonstrates improved performance compared to a model trained solely on a real dataset (\ie, Real-only).
Even the approaches that utilize pseudo-label annotations enhance the generalization performance of TPE trained only with real data.
This result highlights PoseSyn's potential for leveraging a broader range of unlabeled images, which could significantly expand the diversity of human appearances and challenging poses.

Furthermore, the performance improvements observed in 2D pose pseudo-labeling approaches are superior to those in 3D pose pseudo-labeling.
This discrepancy arises due to the added complexity of 3D pose pseudo-labeling, which considers depth in three-dimensional space, potentially reducing accuracy when identifying problematic poses in the EEM framework.

\begin{figure*}[b]
\begin{center}
\centerline{\includegraphics[width=1\linewidth]
{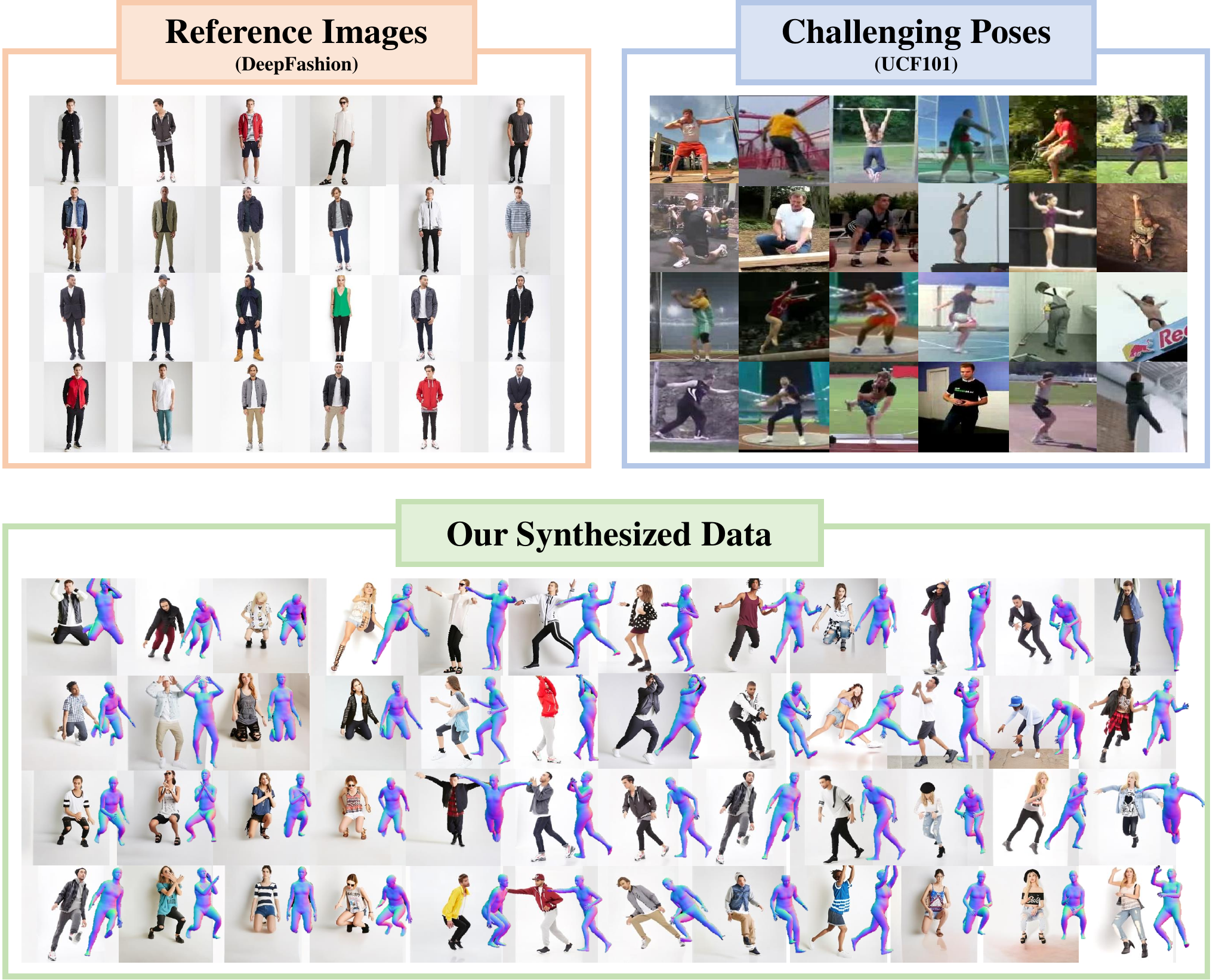}}
\vskip -0.05in
\caption{\textbf{Our synthesized Data.}
We demonstrate the extendibility of our approach in augmenting any easily accessible in-the-wild unlabeled images into a 3D pose dataset.
Here, we incorporate DeepFashion dataset (as reference images) and UCF101 dataset (as challenging poses) to create our synthesized 3D pose dataset.
}
\label{fig:fasuon_ucf}
\end{center}
\vskip -0.2in
\end{figure*}

Finally, to demonstrate that our method can even leverage various datasets beyond the MPII dataset, we showcase our synthesized dataset which incorporates DeepFashion~\cite{liuLQWTcvpr16DeepFashion} dataset for reference images and UCF101~\cite{DBLP:journals/corr/abs-1212-0402} dataset for challenging poses, as shown in Fig.~\ref{fig:fasuon_ucf}.

The DeepFashion dataset features humans with diverse appearances, making it an excellent choice for reference images.
On the other hand, the UCF101 dataset consists of various in-the-wild human poses, making it attractive for use as challenging poses.

\clearpage
\newpage

\section{More Results}\label{more_results}

\paragraph{Reference Images in Motion-guided Video Generation}

\begin{figure}[ht]
    \centering
    \begin{subfigure}[b]{\textwidth}
        \centering
        \includegraphics[width=0.8\linewidth]{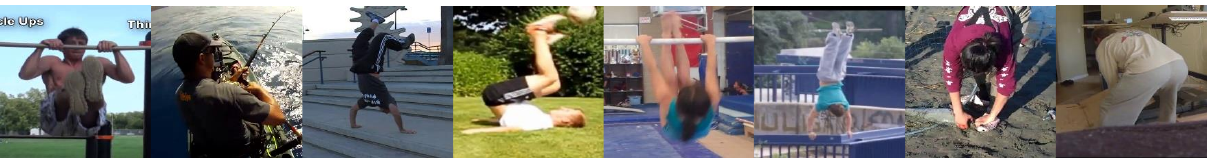}
        \caption{Challenging images}
    \end{subfigure}
    
    \vspace{0.5em} 
    
    \begin{subfigure}[b]{\textwidth}
        \centering
        \includegraphics[width=0.8\linewidth]{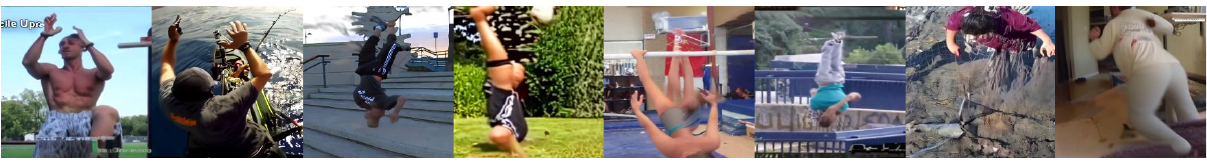}
        \caption{Generated images}
    \end{subfigure}

\caption{\textbf{Experiments on Reference Image in Motion-guided Video Generation.}
The first row displays the challenging images identified through EEM, while the second row illustrates the generated images when these challenging images are served as reference images for the human animation model.
}
\label{fig:challenging_image}
\end{figure}

We utilize the non-challenging images identified through EEM as reference images in the Motion-guided Video Generation stage. 
Using challenging images as reference can hinder the motion-guided generative model's~\cite{zhu2024champ} ability to perform parametric shape alignment in human animation. This misalignment significantly degrades the quality of the animated images, as shown in Fig.~\ref{fig:challenging_image}.

\begin{figure}[h]
    \centering
    \begin{subfigure}[b]{\textwidth}
        \centering
        \includegraphics[width=0.8\linewidth]{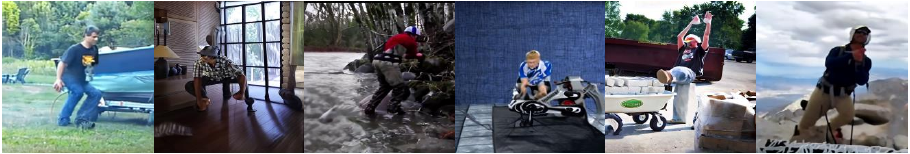}
        \vspace{0.2em}
        \caption{Background blending}
    \end{subfigure}
    
    \vspace{0.7em}
    
    \begin{subfigure}[b]{\textwidth}
        \centering
        \includegraphics[width=0.8\linewidth]{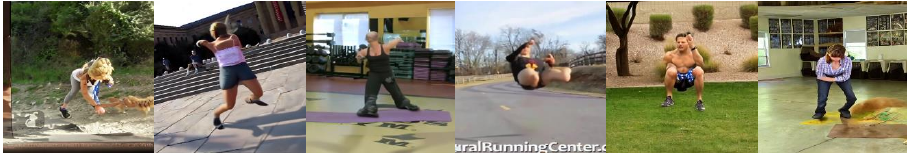}
        \vspace{0.2em}
        \caption{Vanishing limb}
    \end{subfigure}
    
\caption{\textbf{Necessity of Filtering Stage.}
The human animation model occasionally generates images with visual artifacts, including either (a) the unnatural blending of human figures with their backgrounds or (b) disappearance of some joints.
These issues highlight the necessity of the filtering stage in our framework.
All presented images are examples of filtered-out samples during the stage. 
}
\label{fig:filtering}
\end{figure}

\paragraph{Filtering Stage}
Even when using non-challenging images as references, the motion-guided video generation model occasionally leads to images with artifacts, such as blending humans with the background or missing joints.
We present the examples of images removed through our filtering process in Fig.~\ref{fig:filtering}.

\clearpage
\newpage

\begin{figure*}[h]
\begin{center}
\centerline{\includegraphics[width=0.85\linewidth]
{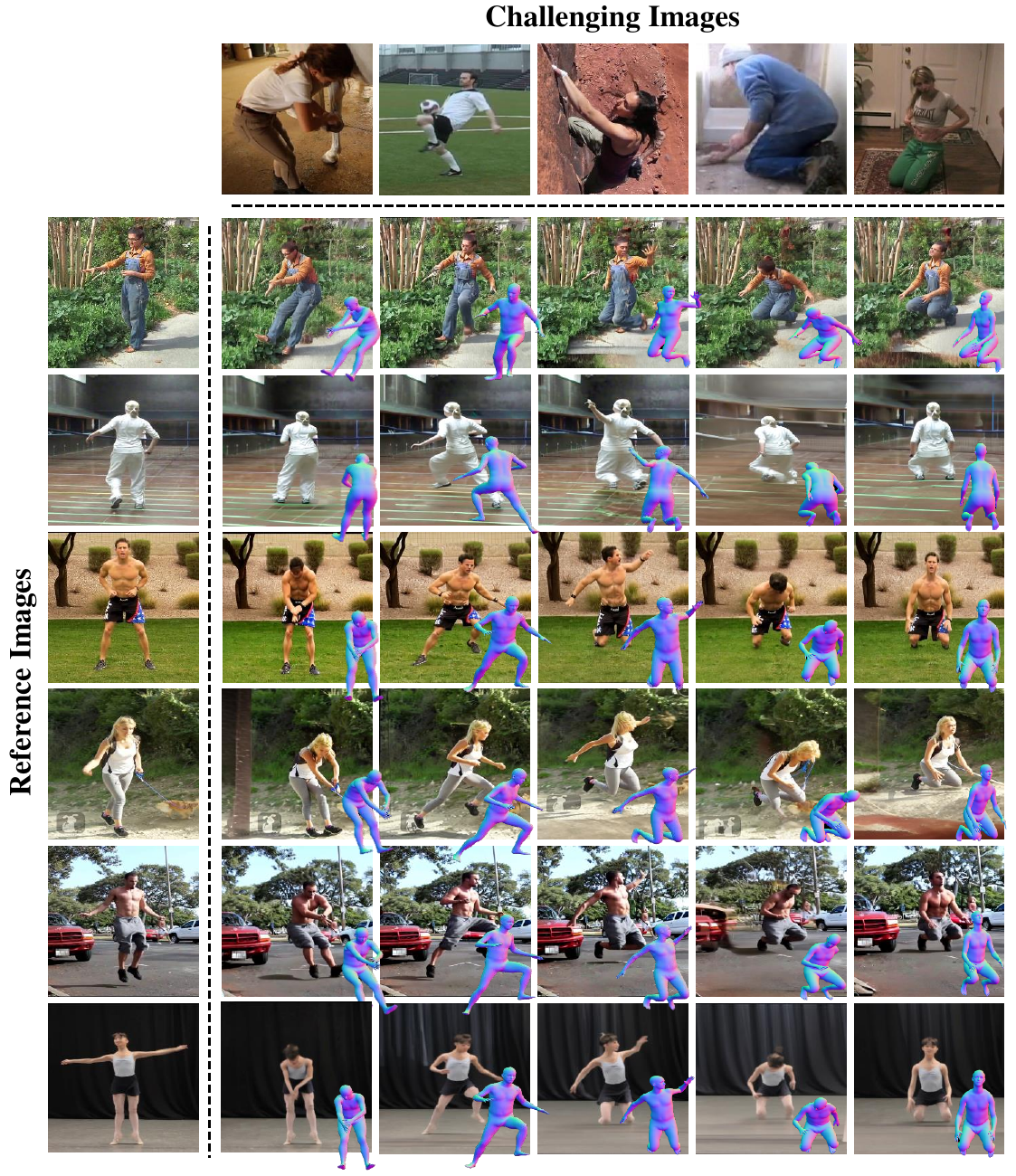}}
\vskip -0.05in
\caption{\textbf{Visualization of Our Synthesized Dataset with Various Combinations.}
The first column on the far left displays non-challenging images identified through EEM, which are served as reference images.
On the other hand, the top row presents challenging images with problematic poses, also identified through EEM, where these poses are augmented into challenging motion sequences via MSM.
The remaining images are synthesized data samples created by combining the non-challenging images with the challenging poses, resulting in diverse datasets with varied appearances, poses, and backgrounds.
}
\label{fig:our_aug1}
\end{center}
\vskip -0.2in
\end{figure*}

\begin{figure*}[h]
\begin{center}
\centerline{\includegraphics[width=0.8\linewidth]
{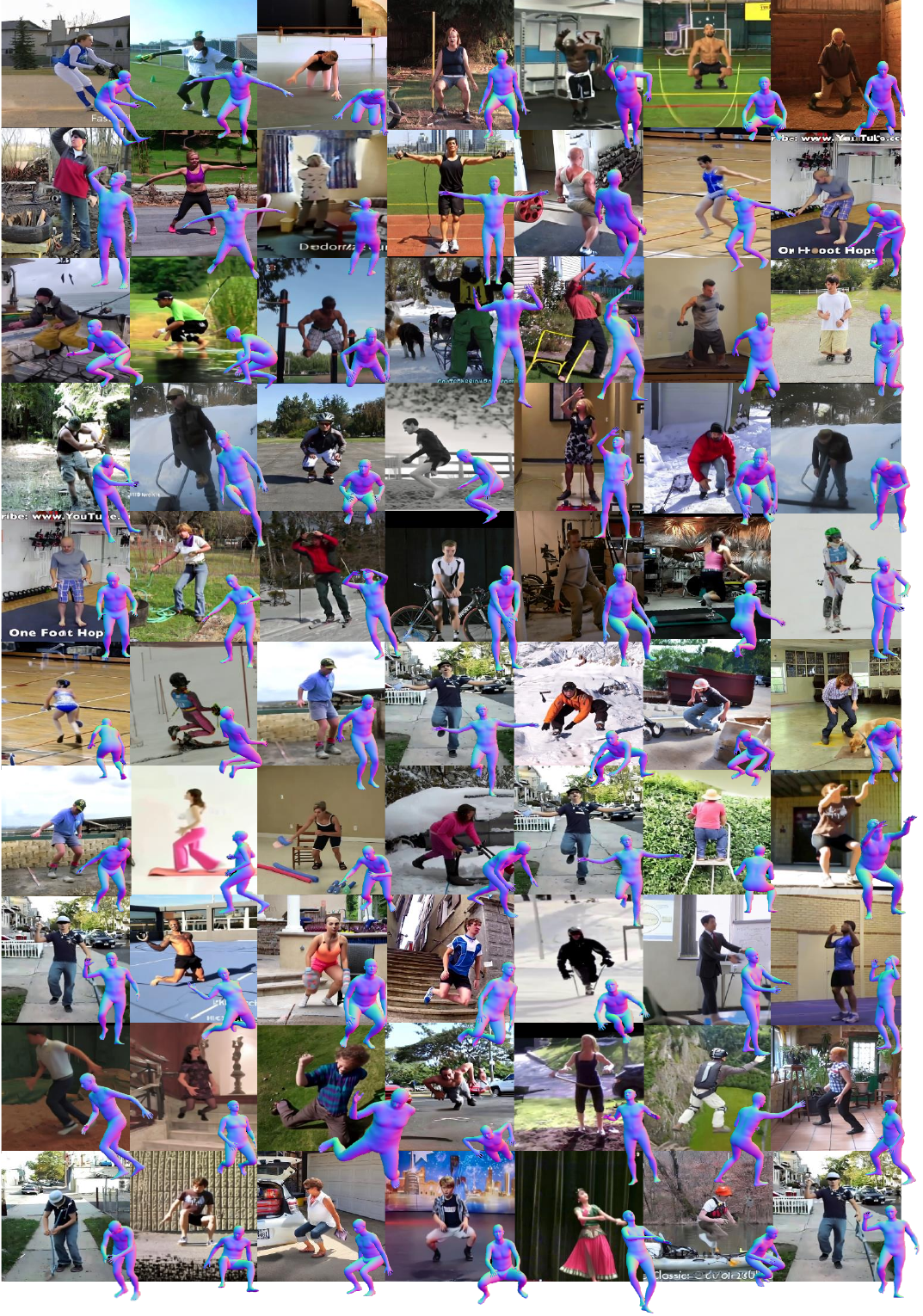}}
\vskip -0.05in
\caption{\textbf{Visualization of Our Synthesized Dataset.}
Our framework synthesizes dataset consisting of highly diverse image sets, featuring various backgrounds, human appearances, and challenging poses.
}
\label{fig:our_aug2}
\end{center}
\vskip -0.2in
\end{figure*}

\paragraph{Diversity in Synthesis}
Our method offers promising advantage of augmenting a single reference image through various challenging poses, as well as augmenting a single problematic pose through various reference images with different viewpoints, human appearances, and backgrounds.
This dual capability allows for the creation of a highly diverse dataset.
We visualize the synthesized data samples in Fig.~\ref{fig:our_aug1} and Fig.~\ref{fig:our_aug2}.

\clearpage
\newpage

\clearpage
\newpage

\section{Additional Qualitative Results}\label{qualitative results}
\begin{figure*}[h]
\begin{center}
\centerline{\includegraphics[width=\linewidth]{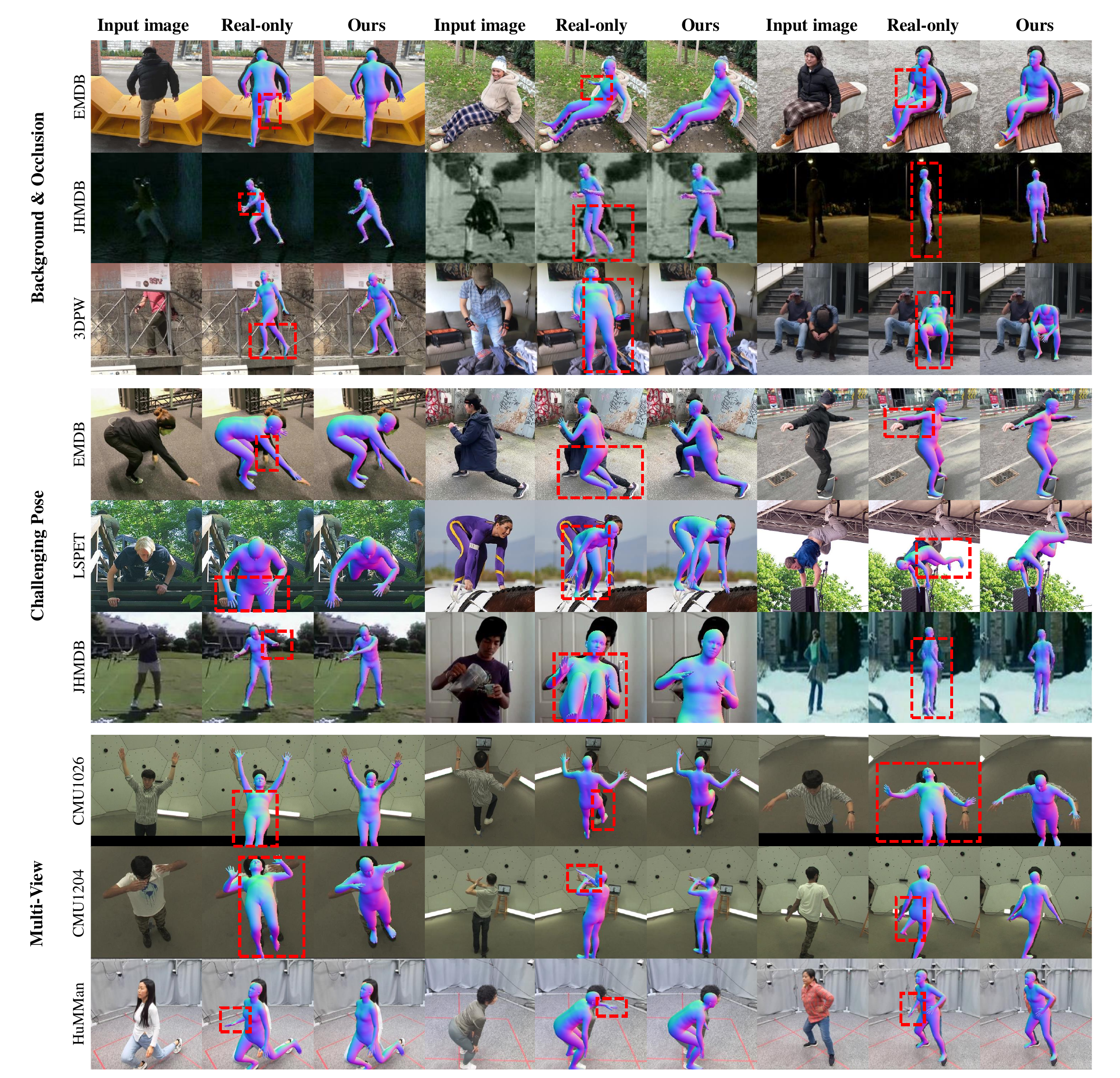}}
\vskip -0.05in
\caption{
\textbf{Qualitative Results (3DCrowdNet).} 
Our approach improves generalization of 3DCrowdNet trained solely on real dataset (\ie, Real-only) across various benchmarks.
Red boxes highlight areas of incorrect predictions in the model trained with real-only data.
}

\label{fig:supp_qualitative_result_3DCrowdNet}
\end{center}
\end{figure*}

\begin{figure*}[h]
\begin{center}
\centerline{\includegraphics[width=\linewidth]{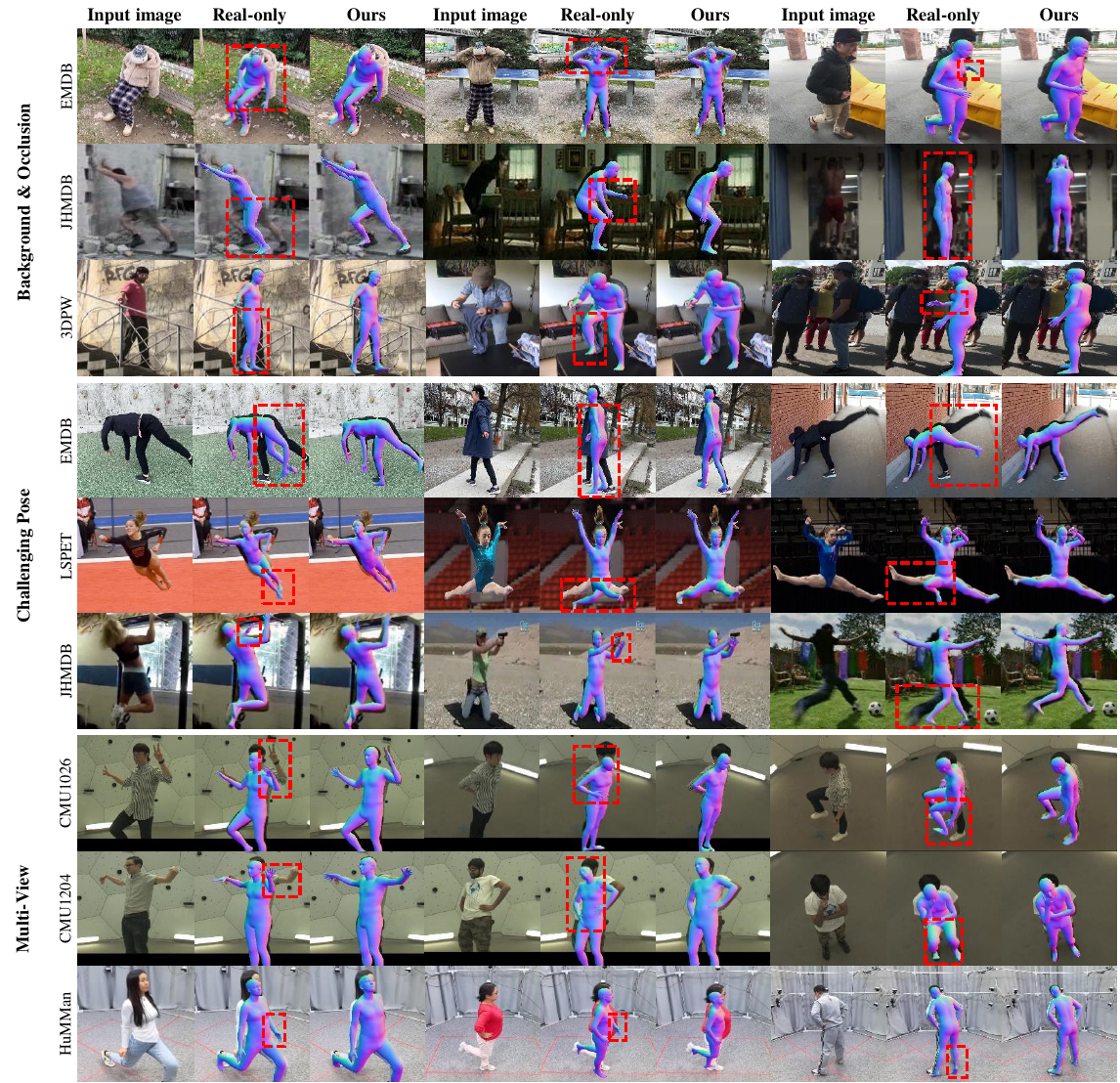}}
\vskip -0.05in
\caption{
\textbf{Qualitative Results (Hybrik).} 
Our approach improves generalization of Hybrik trained solely on real dataset (\ie, Real-only) across various benchmarks.
Red boxes highlight areas of incorrect predictions in the model trained with real-only data.
}
\label{fig:supp_qualitative_result_hybrik}
\end{center}
\end{figure*}

\begin{figure*}[h]
\begin{center}
\centerline{\includegraphics[width=\linewidth]{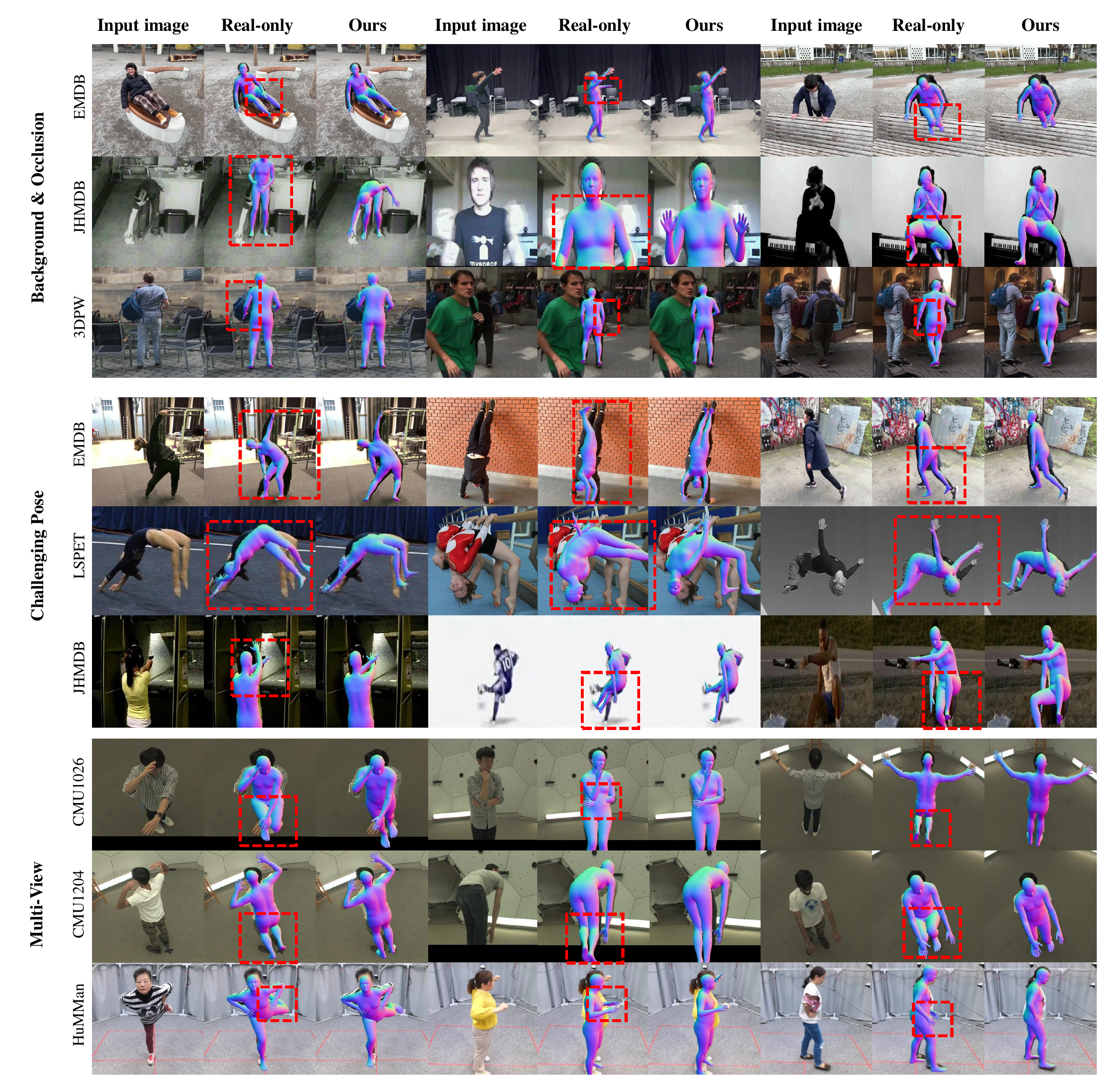}}
\vskip -0.05in
\caption{
\textbf{Qualitative Results (4DHumans).} 
Our approach improves generalization of 4DHumans trained solely on real dataset (\ie, Real-only) across various benchmarks.
Red boxes highlight areas of incorrect predictions in the model trained with real-only data.
}
\label{fig:supp_qualitative_result_vanilla}
\end{center}
\end{figure*}

We present additional qualitative comparison results using three types of TPEs (\ie, 3DCrowdNet, Hybrik, and 4DHumans) in Fig.~\ref{fig:supp_qualitative_result_3DCrowdNet}, Fig.~\ref{fig:supp_qualitative_result_hybrik}, and Fig.~\ref{fig:supp_qualitative_result_vanilla}.

% \end{document}

%% file: folder_tex/Notation_Table.tex
\begin{table*}[t]

    \resizebox{\linewidth}{!}{
    \begin{tabular}{cccc}
    \hline
    
    \multicolumn{1}{c|}{\cellcolor[HTML]{EFEFEF}\textbf{Symbol}}  & \multicolumn{1}{c|}{\cellcolor[HTML]{EFEFEF}\textbf{Description}}       & \multicolumn{1}{c|}{\cellcolor[HTML]{EFEFEF}\textbf{dimension}} & \cellcolor[HTML]{EFEFEF}\textbf{Value} \\ \hline\hline
    
    \multicolumn{4}{c}{\cellcolor[HTML]{EFEFEF}\textbf{Notation related to dataset}}   \\ \hline
    
    \multicolumn{1}{c|}{\cellcolor[HTML]{FFFFFF}$\mathcal{D}$} & \multicolumn{1}{l|}{\cellcolor[HTML]{FFFFFF}2D Pose Dataset. In our experiments, we used MPII dataset. }  & \multicolumn{1}{c|}{\cellcolor[HTML]{FFFFFF}-}        & -         \\ 
    
    \multicolumn{1}{c|}{\cellcolor[HTML]{FFFFFF}$\mathcal{D}_{\text{C}}$}   & \multicolumn{1}{l|}{\cellcolor[HTML]{FFFFFF}A set of challenging dataset  identified through EEM.}       & \multicolumn{1}{c|}{\cellcolor[HTML]{FFFFFF}-}        & -         \\ 
    
    \multicolumn{1}{c|}{\cellcolor[HTML]{FFFFFF}$\mathcal{D}_{\text{NC}}$}  & \multicolumn{1}{l|}{\cellcolor[HTML]{FFFFFF}A set of non-challenging dataset identified through EEM.}   & \multicolumn{1}{c|}{\cellcolor[HTML]{FFFFFF}-}        & -         \\ \hline
    
    \multicolumn{4}{c}{\cellcolor[HTML]{EFEFEF} \textbf{Notation related to image}}     \\ \hline
    
    \multicolumn{1}{c|}{\cellcolor[HTML]{FFFFFF}$I$} & \multicolumn{1}{l|}{\cellcolor[HTML]{FFFFFF}Input image}      & \multicolumn{1}{c|}{\cellcolor[HTML]{FFFFFF}(256, 256, 3)}        & -         \\ 
    \multicolumn{1}{c|}{\cellcolor[HTML]{FFFFFF}$W$} & \multicolumn{1}{l|}{\cellcolor[HTML]{FFFFFF}Image width}      & \multicolumn{1}{c|}{\cellcolor[HTML]{FFFFFF}-}        & -         \\ 
    \multicolumn{1}{c|}{\cellcolor[HTML]{FFFFFF}$H$} & \multicolumn{1}{l|}{\cellcolor[HTML]{FFFFFF}Image height}      & \multicolumn{1}{c|}{\cellcolor[HTML]{FFFFFF}-}        & -         \\ 
    
    \multicolumn{1}{c|}{\cellcolor[HTML]{FFFFFF}$I_{\text{C}}$}   & \multicolumn{1}{l|}{\cellcolor[HTML]{FFFFFF} Challenging image} & \multicolumn{1}{c|}{\cellcolor[HTML]{FFFFFF}(256, 256, 3)}        & -         \\ 
    
    \multicolumn{1}{c|}{\cellcolor[HTML]{FFFFFF}$I_{\text{NC}}$}  & \multicolumn{1}{l|}{\cellcolor[HTML]{FFFFFF}Non-challenging image}      & \multicolumn{1}{c|}{\cellcolor[HTML]{FFFFFF}(256, 256, 3)}        & -         \\ \hline
    
    \multicolumn{4}{c}{\cellcolor[HTML]{EFEFEF}\textbf{Notation related to joint}}        \\ \hline
    
    \multicolumn{1}{c|}{\cellcolor[HTML]{FFFFFF}$N_{\text{2D}}$}  & \multicolumn{1}{l|}{\cellcolor[HTML]{FFFFFF}The number of 2D joints.}  & \multicolumn{1}{c|}{\cellcolor[HTML]{FFFFFF}-}   & 16         \\ 
    \multicolumn{1}{c|}{\cellcolor[HTML]{FFFFFF}$N_{\text{3D}}$}  & \multicolumn{1}{l|}{\cellcolor[HTML]{FFFFFF}The number of 3D joints.}  & \multicolumn{1}{c|}{\cellcolor[HTML]{FFFFFF}-}   & 24         \\ 
    \multicolumn{1}{c|}{\cellcolor[HTML]{FFFFFF}$J^{\text{2D}}_{\text{GT}}$}  & \multicolumn{1}{l|}{\cellcolor[HTML]{FFFFFF}Ground truth 2D pose.}  & \multicolumn{1}{c|}{\cellcolor[HTML]{FFFFFF}(16, 2)}   & -         \\ 
     
        \multicolumn{1}{c|}{\cellcolor[HTML]{FFFFFF}$\hat{J}^{\text{2D}}$}     & \multicolumn{1}{l|}{\cellcolor[HTML]{FFFFFF}2D projection of the 3D pose predicted on the input image through TPE.}   & \multicolumn{1}{c|}{\cellcolor[HTML]{FFFFFF}(16, 2)}   & -         \\ 
    
    \multicolumn{1}{c|}{\cellcolor[HTML]{FFFFFF}$\hat{J}^{\text{3D}}$}     & \multicolumn{1}{l|}{\cellcolor[HTML]{FFFFFF}3D pose predicted on the input image through TPE.}         & \multicolumn{1}{c|}{\cellcolor[HTML]{FFFFFF}(24, 3)}   & -         \\ 
    
    \multicolumn{1}{c|}{\cellcolor[HTML]{FFFFFF}$\hat{J}^{\text{3D}}_{\text{C}}$}  & \multicolumn{1}{l|}{\cellcolor[HTML]{FFFFFF}3D pose predicted on the challenging image through TPE.}     & \multicolumn{1}{c|}{\cellcolor[HTML]{FFFFFF}(24, 3)}   & -         \\ 
    
    \multicolumn{1}{c|}{\cellcolor[HTML]{FFFFFF}$\hat{J}^{\text{3D}}_{\text{NC}}$} & \multicolumn{1}{l|}{\cellcolor[HTML]{FFFFFF}3D pose predicted on non-challenging image through TPE.}  & \multicolumn{1}{c|}{\cellcolor[HTML]{FFFFFF}(24, 3)}   & -         \\ 
    
    \multicolumn{1}{c|}{\cellcolor[HTML]{FFFFFF}$\hat{J}^{\text{2D},n}$}    & \multicolumn{1}{l|}{\cellcolor[HTML]{FFFFFF}Body part index values of 2D pose (\eg, right shoulder, right leg, left shoulder, etc.).}         & \multicolumn{1}{c|}{\cellcolor[HTML]{FFFFFF}(2)}      & -         \\ 
    
    \multicolumn{1}{c|}{\cellcolor[HTML]{FFFFFF}$\hat{J}^{\text{3D},n}$}    & \multicolumn{1}{l|}{\cellcolor[HTML]{FFFFFF}Body part index values of 3D pose (\eg, right shoulder, right leg, left shoulder, etc.).}         & \multicolumn{1}{c|}{\cellcolor[HTML]{FFFFFF}(2)}      & -         \\ \hline
    
    \multicolumn{4}{c}{\cellcolor[HTML]{EFEFEF}\textbf{Notation related to EEM}}        \\ \hline
    
    \multicolumn{1}{c|}{\cellcolor[HTML]{FFFFFF}$Err$}    & \multicolumn{1}{l|}{\cellcolor[HTML]{FFFFFF}Error value calculated by Eq.~(1) to identify challenging dataset.} & \multicolumn{1}{c|}{\cellcolor[HTML]{FFFFFF} - }      & -         \\ 
    \multicolumn{1}{c|}{\cellcolor[HTML]{FFFFFF}$\mathbf{w}_n$}   & \multicolumn{1}{l|}{\cellcolor[HTML]{FFFFFF}Hyperparameter considering the importance of joint parts in $Err$ value calculation.}    & \multicolumn{1}{c|}{\cellcolor[HTML]{FFFFFF}(16)}     & \begin{tabular}[c]{@{}c@{}@{}}Ankle(1), wrist(1), \\ Elbow(0.5), knee(0.5), \\ Hip(0.25), shoulder(0.25)\end{tabular} \\ 
    
    \multicolumn{1}{c|}{\cellcolor[HTML]{FFFFFF}$\text{Top}_{K_{\text{C}}}$}         & \multicolumn{1}{l|}{\cellcolor[HTML]{FFFFFF}Operation used to identify $K_{\text{C}}$ challenging dataset $\mathcal{D}_{\text{C}}$.}      & \multicolumn{1}{c|}{\cellcolor[HTML]{FFFFFF} - }        & 500       \\ 
    
    \multicolumn{1}{c|}{\cellcolor[HTML]{FFFFFF}$\text{Top}_{K_\text{NC}}$}        & \multicolumn{1}{l|}{\cellcolor[HTML]{FFFFFF}Operation used to identify $K_{\text{NC}}$ non-challenging dataset $\mathcal{D}_{\text{NC}}$.}     & \multicolumn{1}{c|}{\cellcolor[HTML]{FFFFFF} - }        & 200       \\ \hline
    
    \multicolumn{4}{c}{\cellcolor[HTML]{EFEFEF}\textbf{Notation related to MSM}}       \\ \hline
    
    \multicolumn{1}{c|}{\cellcolor[HTML]{FFFFFF}$\mathcal{MR}_{\text{init}}$} & \multicolumn{1}{l|}{\cellcolor[HTML]{FFFFFF}Initial motion representation obtained by repeating mis-predicted pose over time step $T$.}    & \multicolumn{1}{c|}{\cellcolor[HTML]{FFFFFF}(30, 263)}  & -         \\ 
    
    \multicolumn{1}{c|}{\cellcolor[HTML]{FFFFFF}\textbf{$\mathcal{Z}_{{\mathcal{MR}}}$}}  & \multicolumn{1}{l|}{\cellcolor[HTML]{FFFFFF}Latent features of initial motion sequences obtained from encoder in VQ-VAE.}        & \multicolumn{1}{c|}{\cellcolor[HTML]{FFFFFF}(7, 512)}  & -         \\ 
    
    \multicolumn{1}{c|}{\cellcolor[HTML]{FFFFFF}$\mathcal{S}_{\mathcal{MR}}$}  & \multicolumn{1}{l|}{\cellcolor[HTML]{FFFFFF}Index values of initial motion sequences obtained through codebook quantization.}    & \multicolumn{1}{c|}{\cellcolor[HTML]{FFFFFF}(7)}      & -         \\ 
    
    \multicolumn{1}{c|}{\cellcolor[HTML]{FFFFFF}$F$} & \multicolumn{1}{l|}{\cellcolor[HTML]{FFFFFF}Preprocessing method~\cite{Guo_2022_CVPR} for motion representation.}      & \multicolumn{1}{c|}{\cellcolor[HTML]{FFFFFF}-}        & -         \\ 
    
    \multicolumn{1}{c|}{\cellcolor[HTML]{FFFFFF}${T}$} & \multicolumn{1}{l|}{\cellcolor[HTML]{FFFFFF}Time step for acquiring a initial motion representation from a initial pose.}      & \multicolumn{1}{c|}{\cellcolor[HTML]{FFFFFF}-}        & 30        \\ 
    
    \multicolumn{1}{c|}{\cellcolor[HTML]{FFFFFF}$\mathcal{E}$} & \multicolumn{1}{l|}{\cellcolor[HTML]{FFFFFF}Encoder of Motion VQ-VAE in T2M-GPT~\cite{zhang2023generating}.}   & \multicolumn{1}{c|}{\cellcolor[HTML]{FFFFFF}-}        & -         \\ 
    
    \multicolumn{1}{c|}{\cellcolor[HTML]{FFFFFF}$M$} & \multicolumn{1}{l|}{\cellcolor[HTML]{FFFFFF}Length of initial motion sequences obtained by dividing $T$ by $r$.}  & \multicolumn{1}{c|}{\cellcolor[HTML]{FFFFFF}-}        & 7         \\ 
    
    \multicolumn{1}{c|}{\cellcolor[HTML]{FFFFFF}$r$} & \multicolumn{1}{l|}{\cellcolor[HTML]{FFFFFF}Temporal downsampling factor of Encoder $\mathcal{E}$.}  & \multicolumn{1}{c|}{\cellcolor[HTML]{FFFFFF}-}        & 4         \\ 
    
    \multicolumn{1}{c|}{\cellcolor[HTML]{FFFFFF}$\mathbf{e}_{\text{text}}$}         & \multicolumn{1}{l|}{\cellcolor[HTML]{FFFFFF}Text embedding of textual description processed by CLIP encoder~\cite{radford2021learning}.}   & \multicolumn{1}{c|}{\cellcolor[HTML]{FFFFFF}(512)}         &  - \\ 
    
    \multicolumn{1}{c|}{\cellcolor[HTML]{FFFFFF}$L$}   & \multicolumn{1}{l|}{\cellcolor[HTML]{FFFFFF}Length of augmented challenging motion sequence $\mathcal{M}_{\text{C}}$.  }     & \multicolumn{1}{c|}{\cellcolor[HTML]{FFFFFF}-}   & -         \\ 
    \multicolumn{1}{c|}{\cellcolor[HTML]{FFFFFF}$\mathcal{M}_{\text{C}}$}   & \multicolumn{1}{l|}{\cellcolor[HTML]{FFFFFF}Augmented challenging motion sequence obtained through transformer under condition of both $\mathbf{e}_{\text{text}}$ and $\mathcal{S}_{\mathcal{MR}}$.  }     & \multicolumn{1}{c|}{\cellcolor[HTML]{FFFFFF}($L$, 22)}   & -         \\ 
    
    \multicolumn{1}{c|}{\cellcolor[HTML]{FFFFFF}$\mathcal{V}_{\text{C}}$}   & \multicolumn{1}{l|}{\cellcolor[HTML]{FFFFFF}Human animated video obtained through~\cite{zhu2024champ} with motion guidance $\mathcal{M}_{\text{C}}$ and reference image $I_{NC}$. }  & \multicolumn{1}{c|}{\cellcolor[HTML]{FFFFFF}($L$, 256, 256, 3)}      & -         \\ \hline
    
    \multicolumn{4}{c}{\cellcolor[HTML]{EFEFEF}\textbf{Notation related to Error \& Filtering}}         \\ \hline
    
    \multicolumn{1}{c|}{\cellcolor[HTML]{FFFFFF}$Err_{3D,l}$}       & \multicolumn{1}{l|}{\cellcolor[HTML]{FFFFFF}Error value calculated by Eq.~(6) in filtering stage.}    & \multicolumn{1}{c|}{\cellcolor[HTML]{FFFFFF}-}        & -       \\ 
    \multicolumn{1}{c|}{\cellcolor[HTML]{FFFFFF}$\tau$}       & \multicolumn{1}{l|}{\cellcolor[HTML]{FFFFFF}Threshold value used in filtering stage.}    & \multicolumn{1}{c|}{\cellcolor[HTML]{FFFFFF}-}        & 120       \\   \hline
    
    \multicolumn{4}{c}{\cellcolor[HTML]{EFEFEF}\textbf{Remaining notations}}   \\ \hline
    
    \multicolumn{1}{c|}{\cellcolor[HTML]{FFFFFF}TPE}    & \multicolumn{1}{l|}{\cellcolor[HTML]{FFFFFF}Top-down human pose estimation model (\eg, 3DCrowdNet~\cite{choi2022learning}, Hybrik~\cite{li2021hybrik}, and 4DHumans~\cite{goel2023humans}).}    & \multicolumn{1}{c|}{\cellcolor[HTML]{FFFFFF}-}        & -         \\ 
    
    \multicolumn{1}{c|}{\cellcolor[HTML]{FFFFFF}f} & \multicolumn{1}{l|}{\cellcolor[HTML]{FFFFFF}Focal length used in 2D projection of 3D pose.}     & \multicolumn{1}{c|}{\cellcolor[HTML]{FFFFFF}(2)}      & \begin{tabular}[c]{@{}c@{}@{}}3DCrwodNet(5000),  4DHumans(5000 / 256 $\times$ $W$), \\ Hybrik(1000 / 256 $\times$ $W$) \end{tabular} \\ 
    
    \multicolumn{1}{c|}{\cellcolor[HTML]{FFFFFF}p}    & \multicolumn{1}{l|}{\cellcolor[HTML]{FFFFFF}Principal point used in 2D projection of 3D pose.}  & \multicolumn{1}{c|}{\cellcolor[HTML]{FFFFFF}(2)}      & Bounding box center    \\ \hline
    \end{tabular}
    }
    \vskip -0.05in
    \caption{\textbf{Notation Table.}
    }
    \vskip -0.1in
    \label{tab:notation}
    \end{table*}
    